\documentclass[a4paper]{article}

\usepackage[english]{babel}
\usepackage[utf8x]{inputenc}
\usepackage[T1]{fontenc}

\usepackage[a4paper,top=3cm,bottom=2cm,left=3cm,right=3cm,marginparwidth=1.75cm]{geometry}

\usepackage{amsfonts}
\usepackage{amsmath}
\usepackage{graphicx}
\usepackage[colorinlistoftodos]{todonotes}
\usepackage[colorlinks=true, allcolors=blue]{hyperref}

\usepackage{multirow}
\usepackage{booktabs}  
\usepackage{comment}
\usepackage{xspace}
\usepackage{flexisym}
\usepackage{booktabs} 
\usepackage{filecontents} 
\usepackage{fancyhdr,tikz}
\usepackage{color}
\usepackage{caption,subcaption}
\usepackage{pgfplots}
\usepackage{pgfplotstable}
\usepackage{algorithmicx}
\usepackage{algorithm}
\usepackage{algpseudocode}
\usepackage{textcomp}
\usepackage{siunitx}

\algnewcommand\algorithmicinput{\textbf{Input:}}
\algnewcommand\INPUT{\item[\algorithmicinput]}
\algnewcommand\algorithmicoutput{\textbf{Output:}}
\algnewcommand\OUTPUT{\item[\algorithmicoutput]}
 \pgfplotsset{every axis  legend/.append  style={
    at={(0,-0.18)},
    anchor=north west}}
\definecolor{bblue}{HTML}{4F81BD}
\definecolor{rred}{HTML}{C0504D}
\definecolor{ggreen}{HTML}{9BBB59}
\definecolor{ppurple}{HTML}{9F4C7C}
\definecolor{yyellow}{HTML}{FFFF00}
\newcommand{\overbar}[1]{\mkern 1.5mu\overline{\mkern-1.5mu#1\mkern-1.5mu}\mkern 1.5mu}

\newcommand{\T}[1]{\ensuremath{\mathcal{#1}}} 
\newcommand{\M}[1]{\ensuremath{\bm{#1}}} 
\newcommand{\V}[1]{\ensuremath{\bm{#1}}} 

\usepackage{xspace,csquotes}
\newcommand{\methodName}{\texttt{COPA}\xspace}
\newcommand{\mname}{\texttt{COPA}\xspace}
\DeclareMathSizes{5}{5}{5}{5}
\usepackage{bm}

\usepackage[font={small,it}]{caption}

\pgfplotsset{
every tick label/.append style={scale=1.1},
}

\begin{document}
\title{ \methodName: Constrained PARAFAC2 for Sparse \& Large  Datasets}

\author{ Ardavan Afshar$^1$, Ioakeim Perros$^1$, Evangelos E. Papalexakis$^2$ \\ Elizabeth Searles$^3$, Joyce Ho$^4$, Jimeng Sun$^1$\\
$^1$Georgia Institute of Technology, $^2$University of California, Riverside\\
$^3$Children's Healthcare Of Atlanta, $^4$Emory University \\
}

\maketitle

\begin{abstract}
PARAFAC2 has demonstrated success in modeling irregular tensors, where the tensor dimensions vary across one of the modes. An example scenario is modeling treatments across a set of patients with the varying number of medical encounters over time.
Despite recent improvements on unconstrained PARAFAC2, its model factors are usually dense and sensitive to noise which limits their interpretability. As a result, the following open challenges remain: 
a) various modeling constraints, such as temporal smoothness, sparsity and non-negativity, are needed to be imposed for interpretable temporal modeling and b) a scalable approach is required to support those constraints efficiently for large datasets.

To tackle these challenges, we propose a {\it CO}nstrained {\it PA}RAFAC2 (\methodName) method, which carefully incorporates optimization constraints such as temporal smoothness, sparsity, and non-negativity in the resulting factors. To efficiently support all those constraints, \methodName\ adopts a hybrid optimization framework using alternating optimization and alternating direction method of multiplier (AO-ADMM). 
As evaluated on large electronic health record (EHR) datasets with hundreds of thousands of patients, \mname 
achieves significant speedups (up to $36\times$ faster) over prior PARAFAC2 approaches that only attempt to handle a subset of the constraints that \mname enables. Overall, our method outperforms all the baselines attempting to handle a subset of the constraints in terms of speed, while achieving the same level of accuracy.

Through a case study on temporal phenotyping of medically complex children, we demonstrate how the constraints imposed by \mname reveal concise phenotypes and meaningful temporal profiles of patients. The clinical interpretation of both the phenotypes and the temporal profiles was confirmed by a medical expert.

\end{abstract}

\section{Introduction} 
\begin{figure}[b]
\centering
\includegraphics[scale=0.25]{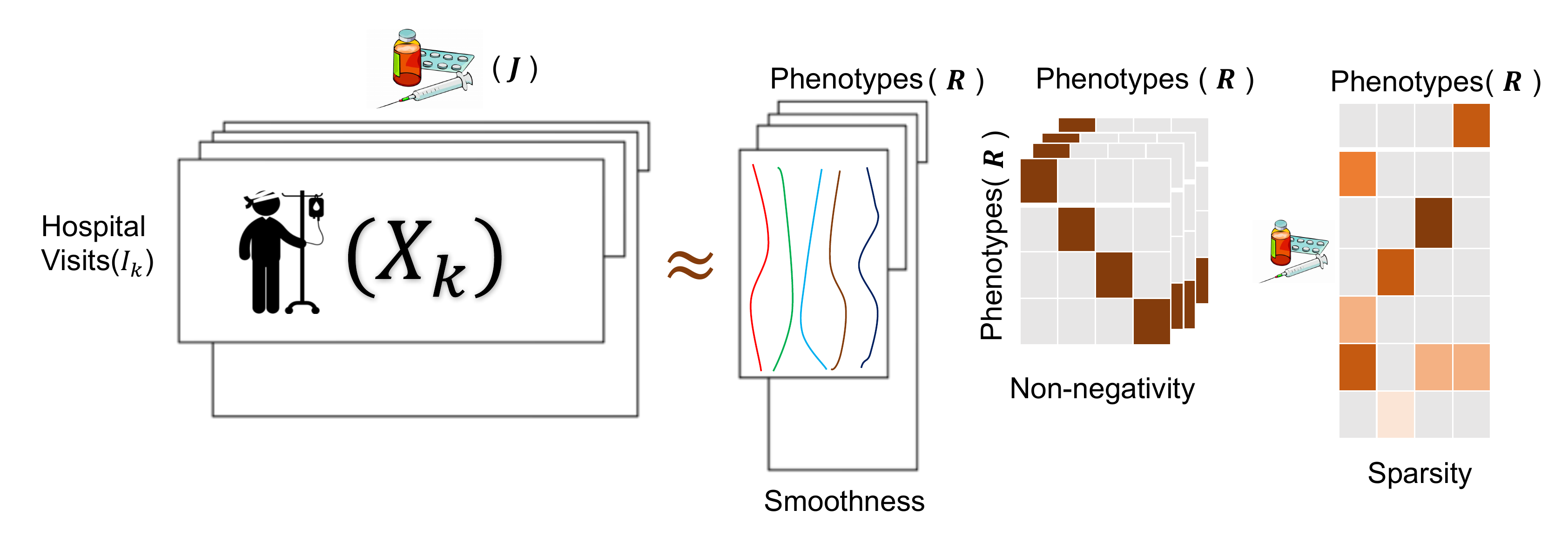}
\caption{\footnotesize An illustration of the constraints imposed by  \mname on PARAFAC2 model factors, targeting temporal phenotyping via EHR data.
}
\label{fig_PARAFAC2}
\end{figure}

Tensor factorization encompasses a set of powerful analytic methods that have been successfully applied in many application domains: social network analysis \cite{Lin2009MetaFacCD,acar2009link}, urban planning \cite{afshar2017cp}, and health analytics \cite{ho2014limestone,ho2014marble,wang2015rubik,Perros2017-dh,perros2018sustain}. 
Despite the recent progression on modeling the time through regular tensor factorization approaches \cite{afshar2017cp, matsubara2014funnel}, there are some cases where modeling the time mode is intrinsically difficult for the regular tensor factorization methods, due to its irregularity. 
A concrete example of such irregularity is electronic health record (EHR). 
 EHR datasets consist of $K$ patients where  patient $k$ is represented using a matrix $\M{X}_k$ and for each patient, $J$ medical features are recorded. Patient $k$ can have $I_k$ hospital visits over time, which can be of different size across patients as shown in Figure ~\ref{fig_PARAFAC2}. 

\begin{table*}[t]
    \centering
     \caption{Comparison of PARAFAC2 models and constrained tensor factorization applied to phenotyping 
     }
    \label{propterty_comp}
\scalebox{0.7}{
\begin{tabular}{ |c|c|c|c|c|c|c|c}
\hline
\parbox[t]{1cm}{\centering Property } &  \parbox[t] {1.9cm}{\centering Marble \cite{ho2014marble} } &  \parbox[t]  {1.9cm}{\centering Rubik \cite{wang2015rubik} } &  \parbox[t]{1.9cm}{ \centering PARAFAC2 \cite{bro1999parafac2}}  & \parbox[t]{1.9cm}{ \centering  SPARTan \cite{Perros2017-dh}} & \parbox[t]{1.7cm}{\centering  Helwig \cite{helwig2017estimating}}& \parbox[t]{1.5cm}{ \centering \methodName}  \\
\hline
Smoothness  &-&- &- & - & \checkmark&\checkmark \\
Sparsity &\checkmark&\checkmark& - & - & -&\checkmark \\
Scalability &-&- &- & \checkmark &  - &\checkmark\\
Handle irregular tensors &-&-& \checkmark & \checkmark &  \checkmark &\checkmark\\
\hline
\end{tabular}}
   
\end{table*}

In this case, clinical visits are the irregular time points which vary across patients. In particular, the {\it time irregularity} lies in 1) the variable number of hospital visits, 2) the varying disease states for different patients, and 3) the varying time gaps between consecutive hospital visits. 
To handle such irregular tensors, the state-of-the-art tensor model is PARAFAC2~\cite{Hars1972b}, which naturally handles variable size along one of its modes (e.g., time mode). 
Despite the wide range of PARAFAC2 applications (e.g., natural language processing~\cite{chew2007cross}, chemical processing~\cite{wise2001application}, and social sciences~\cite{helwig2017estimating}) its computational requirements have limited its usage for small and dense datasets~\cite{kiers1999parafac2}. Even if recently, a scalable PARAFAC2 fitting algorithm was proposed for large, sparse data~\cite{Perros2017-dh}, it cannot incorporate meaningful constraints on the model factors such as: a) sparsity, which facilitates model inspection and understanding and b) smoothness, which is meaningful to impose when temporal evolution is modeled as a mode of the input tensor. 

To tackle the above challenges, we propose  the COnstrained PARAFAC2 method (\mname), which introduces various useful constraints in PARAFAC2 modeling. In particular, generalized temporal smoothness constraints are integrated in order to: a) properly model temporally-evolving phenomena (e.g., evolving disease states), and b) adaptively deal with uneven spacing along the temporal dimension (e.g., when the time duration between consecutive hospital visits may range from $1$ day to several years).
Also, \mname introduces sparsity into the latent factors, a crucial property enhancing interpretability for sparse input data, such as the EHR. 

A key property of our approach is that those constraints are introduced in a computationally efficient manner. To do so, \mname adopts a hybrid optimization framework using alternating optimization and alternating direction method of multipliers. This enables our approach to achieve significant speedups (up to $36\times$) over baselines supporting only a specific constraint each, while achieving the same level of accuracy. Through both quantitative (e.g., the percentage of sparsity) and qualitative evaluations from a clinical expert, we demonstrate the meaningfulness of the constrained output factors for the task of temporal phenotyping via EHRs. In summary, we list our main contributions below:
\begin{itemize}
\item \textbf{Constrained PARAFAC2:} We propose \mname, a method equipping the PARAFAC2 modeling with a variety of meaningful constraints such as smoothness, sparsity, and non-negativity.

\item \textbf{Scalable PARAFAC2:} While \mname incorporates a wide range of constraints, it is faster and more scalable than baselines supporting only a subset of those constraints.

\item \textbf{\mname for temporal phenotyping:} We apply \mname for temporal phenotyping of a medically complex population; a medical expert confirmed the clinical meaningfulness of the extracted phenotypes and temporal patient profiles.

\end{itemize}
Table~\ref{propterty_comp} summarizes the contributions in the context of existing works.

\section{Background}
In this Section, we provide the necessary background for tensor operations.  Then, we briefly illustrate the related work including: the classical method for PARAFAC2 and AO-ADMM framework for constrained tensor factorization. Table \ref{symbol} summarizes the notations used throughout the paper.  
\begin{table}[!ht] 
    \centering
     \caption{\footnotesize Symbols and notations used throughout the paper.}
     \label{symbol}
\scalebox{0.8}{
\begin{tabular}{ |c|c|}
\hline
\parbox[t]{1cm}{\centering Symbol} &  \parbox[t]{4cm}{ \centering Definition }  \\
\hline
*  & Element-wise Multiplication  \\

$\odot$ & Khatri Rao Product \\
$\circ$ & Outer Product \\
\hline
$c(\M{Y})$ & A constraint on factor matrix \M{Y} \\
$\overbar{\M{Y}}$ & Auxiliary variable for factor matrix \M{Y} \\
\hline
$\T{Y},\M{Y},\V{y}$ & Tensor, matrix, vector \\
$\M{Y_{(n)}}$  & Mode-$n$ Matricization of $\T{Y}$ \\
$\V{Y(i,:)}$ & Spans the entire $i$-th row of $\M{Y}$  \\
$\M{X_k}$ & $k^{th}$ frontal slice of tensor $\T{X}$ \\
$diag(\V{y})$ & Diagonal matrix with vector $\V{y}$ on diagonal \\
$diag(\M{Y})$ & Extract the diagonal of matrix $\M{Y}$ \\

\hline
\end{tabular}}
\end{table}

The \textit{mode} or \textit{order} is the number of dimensions of a tensor.
A \textit{slice} refers to a matrix derived from the tensor where fixing all modes but two. 
\textit{Matricization} converts the tensor into a matrix representation without altering its values. The mode-n matricization of $\T{Y} \in \mathbb{R}^{I_1 \times .... \times I_N} $ is denoted as {\small$\M{Y_{(n)}} \in \mathbb{R}^{I_n \times I_1..I_{n-1} I_{n+1}..I_{N}}$}. Matricized-Tensor-Times-Khatri-Rao-Product\cite{bader2007efficient} (MTTKRP) is  a multiplication which a naive construction of that for large and sparse tensors needs computational cost and enormous memory and is the typical bottleneck in most tensor factorization problems. 



\noindent The popular CP decomposition \cite{carroll1970analysis}  also known as PARAFAC factorizes a tensor into a sum of R rank-one tensors. CP decomposition method factorizes tensor $\T{Y} \in \mathbb{R}^{K \times J \times I} $  into $\sum_{r=1}^{R} \V{a_r} \circ \V{b_r} \circ \V{c_r}$ where R is the number of target-ranks or components and $\V{a_r} \in \mathbb{R}^{K}$, $\V{b_r} \in \mathbb{R}^{J}$, and $\V{c_r} \in \mathbb{R}^{I}$ are column matrices and $\circ$ indicates the outer product. Here $\M{A}=[\V{a_1},...\V{a_R}]$, $\M{B}=[\V{b_1},...\V{b_R}]$, and $\M{C}=[\V{c_1},...\V{c_R}]$ are factor matrices.  
\\ \textbf{Original PARAFAC2 model}
As proposed in~\cite{Hars1972b}, the PARAFAC2 model decomposes each slice of the input ${\small\M{X_k}\in \mathbb{R}^{I_k \times J}}$ as {\small$\M{X_k} \approx \M{U_k} \M{S_k} \M{V^T}$}, where {\small$\M{U_k} \in \mathbb{R}^{I_k \times R}$}, {\small$\M{S_k} \in \mathbb{R}^{R \times R}$} is a diagonal matrix, and {\small$\M{V} \in \mathbb{R}^{J \times R}$}. Uniqueness is an important property in factorization models which ensures that the pursued solution is not an arbitrarily rotated version of the actual latent factors. 
In order to enforce uniqueness, Harshman~\cite{Hars1972b} imposed the constraint {\small$\M{U_k^T}\M{U_k} =\Phi \quad \forall k$}. This is equivalent to each {\small$U_k$} being decomposed as {\small$\M{U_k}=\M{Q_K} \M{H}$}, where {\small$\M{Q_k} \in \mathbb{R}^{I_k \times R}$}, {\small$\M{Q_k^T} \M{Q_k}=\M{I} \in \mathbb{R}^{R \times R}$}, and {\small$\M{H} \in \mathbb{R}^{R \times R}$}. Note that {\small$\M{Q_k}$} has orthonormal columns and {\small$\M{H}$} is invariant regardless of {\small$k$}. Therefore, the decomposition of {\small$\M{U_k}$} implicitly enforces the constraint as {\small$\M{U_k^T} \M{U_k} = \M{H^T} \M{Q_k^T} \M{Q_k} \M{H}=\M{H^T} \M{H} = \Phi$}. Given the above modeling, the standard algorithm~\cite{kiers1999parafac2} to fit PARAFAC2 for dense input data tackles the following optimization problem:
\begin{equation}
\small
\begin{aligned}
& \underset{\{\M{U_k}\}, \{\M{S_k}\}, \M{V}}{\text{minimize}}
& & \sum_{k=1}^{K} \frac{1}{2}||\M{X_k} -\M{U_k}\M{S_k}\M{V^T}||_F^2
\end{aligned}
\label{Classic_PARAFAC2_obj_func}
\end{equation}
subject to {\small$\M{U_k}=\M{Q_K} \M{H}, \M{Q_k^T} \M{Q_k}=\M{I}$}, and {\small$\M{S_k}$} is diagonal.
The solution follows an Alternating Least Squares (ALS) approach to update the modes. First, orthogonal matrices {\small$\{\M{Q_k}\}$} are solved by fixing {\small$\M{H}$},{\small$\{\M{S_k}\}$}, {\small$\M{V}$} and posing each {\small$\M{Q_k}$} as an individual Orthogonal Procrustes Problem \cite{schonemann1966generalized}:
\begin{equation}
\small
\begin{aligned}
& \underset{\{\M{Q_k}\}}{\text{minimize}}
& &  \frac{1}{2} ||\M{X_k}- \M{Q_k} \M{H} \M{S_k} \M{V^T}||_F^2 \\
\end{aligned}
\label{solv_Qk}
\end{equation}
Computing the singular value decomposition (SVD) of {\small$\M{H} \M{S_k} \M{V^T} \M{X_k^T} = \M{P_k} \M{\Sigma_k} \M{Z_k}^T$} yields the optimal {\small$\M{Q_k} = \M{P_k} \M{Z_k}^T$}. With {\small$\{\M{Q_k}\}$} fixed, the remaining factors can be solved as:
\begin{equation}
\small
\begin{aligned}
& \underset{\M{H},\{\M{S_k}\},\M{V}}{\text{minimize}}
& &  \frac{1}{2} \sum_{k=1}^{K}||\M{Q_k^T} \M{X_k}-\M{H} \M{S_k} \M{V^T}||_F^2 \\
\end{aligned}
\label{solv_HSV}
\end{equation}
The above is equivalent to the CP decomposition of tensor {\small$\T{Y} \in \mathbb{R}^{R \times J \times K}$} with slices {\small$\M{Y_k} = \M{Q_k^T} \M{X_k}$}. A single iteration of the CP-ALS provides an update for {\small$\M{H}$,$\{\M{S_k}\}$}, {\small$\M{V}$} \cite{kiers1999parafac2}. The algorithm iterates between the two steps (Equations \ref{solv_Qk} and \ref{solv_HSV}) until convergence is reached.

\noindent \textbf{AO-ADMM} Recently, a hybrid algorithmic framework, AO-ADMM \cite{huang2016flexible}, was proposed for constrained CP factorization based on alternating optimization (AO) and the alternating direction method of multipliers (ADMM). Under this approach, each factor matrix is updated iteratively using ADMM while the other factors are fixed. A variety of constraints can be placed on the factor matrices, which can be readily accommodated by ADMM. 

\section{Proposed Method: Constrained PARAFAC2 Framework}
A generalized constrained PARAFAC2 approach is appealing from several perspectives including the ability to encode prior knowledge, improved interpretability, and more robust and reliable results. We propose \methodName, a scalable and generalized constrained PARAFAC2 model, to impose a variety of constraints on the factors. Our algorithm leverages AO-ADMM style iterative updates for some of the factor matrices and introduces several PARAFAC2-specific techniques to improve computational efficiency. Our framework has the following benefits:
\begin{itemize}
\item Multiple constraints can be introduced simultaneously.
\item The ability to handle large data as solving the constraints involves the application of several element-wise operations.
\item Generalized temporal smoothness constraint that effectively deals with uneven spacing along the temporal (irregular) dimension (gaps from a day to several years for two consecutive clinical visits).
\end{itemize}

In this section, we first illustrate the general framework for formulating and solving the constrained PARAFAC2 problem. We then discuss several special constraints that are useful for the application of phenotyping including sparsity on the {\small$\M{V}$}, smoothness on the {\small$\M{U_k}$}, and non-negativity on  the {\small$\M{S_k}$} factor matrices in more detail.

\subsection{General Framework for \mname}
The constrained PARAFAC2 decomposition can be formulated using generalized constraints on $\M{H}$, {\small$\M{S_k}$}, and {\small$\M{V}$}, in the form of {\small$c(\M{H})$}, {\small$c(\M{S_k})$}, and {\small$c(\M{V})$} as: 
\begin{equation*}
\small
\begin{aligned}
& \underset{\{\M{U_k}\}, \{\M{S_k}\}, \M{V}}{\text{minimize}}
& & \sum_{k=1}^{K} \frac{1}{2}||\M{X_k} -\M{U_k}\M{S_k}\M{V^T}||_F^2 +c(\M{H})+\sum_{k=1}^{K} c(\M{S}_k) + c(\M{V})
\end{aligned}
\end{equation*}

subject to {\small$\M{U_k}=\M{Q_K} \M{H}, \M{Q_k^T} \M{Q_k}=\M{I}$}, and {\small$\M{S_k}$} is diagonal.
To solve for those constraints, we introduce auxiliary variables for {\small$\M{H}$}, {\small$\M{S_k}$}, and {\small$\M{V}$} (denoted as {\small$\overbar{\M{H}}$}, {\small$\overbar{\M{S_k}}$}, and {\small$\overbar{\M{V}}$}). 
Thus, the optimization problem has the following form:
\begin{equation}
\small
\begin{aligned}
& \underset{\{\M{U_k}\}, \{\M{S_k}\}, \M{V}}{\text{minimize}}
& & \sum_{k=1}^{K} \frac{1}{2}||\M{X_k} -\M{U_k}\M{S_k}\M{V^T}||_F^2 +c(\overbar{\M{H}})+\sum_{k=1}^{K} c(\overbar{\M{S_k}}) + c(\overbar{\M{V}}) \\
& \text{subject to}
& & \M{U_k}=\M{Q_k} \M{H}, \M{Q_k^T} \M{Q_k}=\M{I}, \M{S_k}=\overbar{\M{S_k}} \text{~for all k=1,...,K} \\
& & & \M{H}=\overbar{\M{H}},  \M{V}=\overbar{\M{V}}
\end{aligned}
\label{PARAFAC2_obj_func}
\end{equation}


\noindent We can re-write the objective function as the minimization of {\small$tr(\M{X_k^T}\M{X_k})-2tr(\M{X_k^T}\M{Q_k}\M{H}\M{S_k}\M{V^T})+tr(\M{V}\M{S_k}\M{H^T}
\M{{Q_k}^T}\M{Q_k}\M{H}\M{S_k}\M{V^T})$} in terms of $\M{Q_k}$. The first term is constant and since  {\small $\M{Q_k}$} has orthonormal columns ($\small \M{{Q_k}^T}\M{Q_k}=\M{I}$), the  third term is also constant. By rearranging the terms we have {\small $tr(\M{X_k^T}\M{Q_k}\M{H}\M{S_k}\M{V^T})=tr(\M{X_k}\M{V}\M{S_k}\M{H^T}\M{Q_k^T})$}.  Thus, the objective function regarding to $\M{Q_k}$ is equivalent to:  
\begin{equation}
\small
\begin{aligned}
& \underset{\M{Q_k}}{\text{minimize}}
& &  \frac{1}{2}||\M{X_k} \M{V} \M{S_k} \M{H^T} - \M{Q_k}||_F^2 \\
& \text{subject to}
&  &  \M{Q_k^T} \M{Q_k}=\M{I}
\end{aligned}
\label{solv_Q}
\end{equation}
Thus, the optimal {\small$\M{Q_k}$} has the closed form solution {\small$\M{Q_k}=\M{B_k} \M{C_k^T}$} where {\small$\M{B_k} \in R^{I_k \times R}$} and {\small$\M{C_k} \in R^{R \times R}$} are the right and left singular vectors of {\small$\M{X_k} \M{V} \M{S_k} \M{H^T}$} \cite{schonemann1966generalized,golub2013matrix}.
This promotes the solution's uniqueness, since orthogonality is essential for uniqueness in the unconstrained case.

Given fixed {\small$\{\M{Q_k}\}$}, we next find the solution for {\small$\M{H}$}, {\small$\{\M{S_k}\}$}, {\small$\M{V}$} as follows:
\begin{equation}
\small
\begin{aligned}
& \underset{\M{H},\{\M{S_k}\},\M{V}}{\text{minimize}}
& &  \frac{1}{2} \sum_{k=1}^{K}||\M{Q_k^T} \M{X_k}-\M{H} \M{S_k} \M{V^T}||_F^2 +c(\overbar{\M{H}})+\sum_{k=1}^{K} c(\overbar{\M{S_k}}) + c(\overbar{\M{V}}) \\
& \text{subject to}
& & \M{S_k}=\overbar{\M{S_k}} \quad \quad \quad \text{for all k=1,...,K} \\
& & & \M{H}=\overbar{\M{H}},   \M{V}=\overbar{\M{V}}
\end{aligned}
\label{solv_others}
\end{equation}This is equivalent to performing a CP decomposition of tensor {\small$\T{Y} \in \mathbb{R}^{R \times J \times K}$} with slices {\small$\M{Y_k} = \M{Q_k^T} \M{X_k}$}~\cite{kiers1999parafac2}. Thus, the objective is of the form:
\begin{equation}
\small
\begin{aligned}
& \underset{\M{H},\M{W},\M{V}}{\text{minimize}}
& &  \frac{1}{2}||\T{Y}-[\M{H};\M{V};\M{W}]||_F^2 + c(\overbar{\M{H}}) +c(\overbar{\M{V}})+ c(\overbar{\M{W}})\\
& & & \M{H}=\overbar{\M{H}}, \M{V}=\overbar{\M{V}}, \M{W}=\overbar{\M{W}}
\end{aligned}
\label{Solve_HWV}
\end{equation}

We use the AO-ADMM approach \cite{huang2016flexible} to compute {\small$\M{H}$}, {\small$\M{V}$}, and {\small$\M{W}$}, where 
{\small$\M{S_k}=diag(\V{W(k,:)})$} and {\small$\M{W} \in \mathbb{R}^{K \times R}$}.
Each factor matrix update is converted to a constrained matrix factorization problem by performing the mode-n matricization of {\small$\M{Y_{(n)}}$} in Equation \ref{Solve_HWV}.
As the updates for {\small$\M{H}$}, {\small$\M{W}$}, and {\small$\M{V}$} take on similar forms, we will illustrate the steps for updating {\small$\M{W}$}.
Thus, the equivalent objective for $\M{W}$ using the 3rd mode matricization of {\small$\T{Y}$} ({\small$\M{Y_{(3)}} \in \mathbb{R}^{K \times RJ}$}) is:

\begin{equation}
\small
\begin{aligned}
& \underset{\M{W^T}}{\text{minimize}}
& &  \frac{1}{2}||\M{Y_{(3)}^T}-(\M{V} \odot \M{H})\M{W^T}||_F^2 + c(\overbar{\M{W}}) \\
& \text{subject to}
& & \M{W^T}=\overbar{\M{W}}
\end{aligned}
\label{Solve_W}
\end{equation}
The application of ADMM yields the following update rules:
\begin{equation*}
\small
\begin{aligned}
\M{W^T} &:=\Big((\M{H^T} \M{H} * \M{V^T} \M{V})+ \rho \M{I}\Big)^{-1}(\M{Y_{(3)}}(\M{V}\odot \M{H})+\rho(\overbar{\M{W}}+\M{D}_{W^T}))^T \\
\overbar{\M{W}}& :=\arg\min_{\overbar{\M{W}}} c(\overbar{\M{W}}) +\frac{\rho}{2} ||\overbar{\M{W}}-\M{W^T}+\M{D}_{W^T}||_F^2 \\
\M{D}_{W^T} &:=\M{D}_{W^T}+\overbar{\M{W}}-\M{W^T}
\end{aligned}
\end{equation*}
where {\small$\M{D}_{W^T}$} is a dual variable  and $\rho$ is a step size regarding to {\small$\M{W^T}$} factor matrix.  The auxiliary variable ({\small$\overbar{\M{W}}$}) update is known as the proximity operator \cite{parikh2014proximal}. 
Parikh and Boyd show that for a wide variety of constraints, the update can be computed using several element-wise operations. 
In Section~\ref{sec:constraints}, we discuss the element-wise operations for three of the constraints we consider.

\subsection{Implementation Optimization}
In this section, we will provide several steps to accelerate the convergence of our algorithm. 
First, our algorithm needs to decompose {\small$\T{Y}$}, therefore, MTTKRP will be a bottleneck for sparse input. Thus \methodName~ uses the fast MTTKRP proposed in SPARTan \cite{Perros2017-dh}. 
Second, {\small$\Big((\M{H^T} \M{H} * \M{V^T} \M{V})+ \rho \M{I}\Big)$} is a symmetric positive definite matrix, therefore instead of calculating the expensive inverse computation, we can calculate the Cholesky decomposition of it ({\small$\M{L} \M{L^T}$}) where {\small$\M{L}$} is a lower triangular matrix and then apply the inverse on {\small$\M{L}$} (lines 14,16 in algorithm \ref{CONSPAR2_alg}). 
Third, {\small$\M{Y_{(3)}}(\M{V}\odot \M{H})$} remains a constant and is unaffected by updates to {\small$\M{W}$} or {\small$\overbar{\M{W}}$}.
Thus, we can cache it and avoid unnecessary re-computations of this value. Fourth,
based on the AO-ADMM results and our own preliminary experiments, our algorithm sets {\small$\rho=\frac{||\M{H^T}\M{H} +\M{V^T}\M{V}||_F^2}{R}$} for fast convergence of {\small$\M{W}$} .

Algorithm \ref{CONSPAR2_alg} lists the pseudocode for solving the generalized constrained PARAFAC2 model.
Adapting AO-ADMM to solve {\small$\M{H}$}, {\small$\M{W}$}, and {\small$\M{V}$} in PARAFAC2 has two benefits: (1) a wide variety of constraints can be incorporated efficiently with iterative updates computed using element-wise operations and (2) computational savings gained by caching the MTTKRP multiplication and using the Cholesky decomposition to calculate the inverse.

\begin{algorithm}
\caption{\methodName}
\begin{algorithmic}[1]

\INPUT  $\M{X_k} \in \mathbb{R}^{I_k\times J} $ for k=1,...,K and target rank R 
\OUTPUT $\M{U_k} \in \mathbb{R}^{I_k \times R}, \M{S_k} \in \mathbb{R}^{R \times R}$  for $k=1,..., K, \M{V} \in \mathbb{R}^{J \times R}$
\Statex
 \\Initialize $\M{H}, \M{V}, \{\M{S_k}\} $ for k=1,...,K 
 \While{convergence criterion is not met}
    
    \For{k=1,...,K }
          \State $[\M{B_k},\M{D_k},\M{C_k}^T]$=truncated SVD of $\M{X_k} \M{V} \M{S_k} \M{H^T}$
          \State $\M{Q_k}=\M{B_k} \M{C_k^T}$   
          \State $\M{Y_k}=\M{Q_k^T} \M{X_k}$
          \State $W(k,:)=diag(\M{S_k})$
    \EndFor
    \State $\M{Z_1}=\M{H}$, $\M{Z_2}=\M{W}$,$\M{Z_3}=\M{V}$
    \For{n=1,...,3 } 
            	\State $\M{G}=*_{i\neq n} \M{Z_i^T}\M{Z_i}$
                \State $\M{F}=\M{Y_{(n)}}( \odot_{i\neq n} \M{Z_i})$  \quad //calculated based on \cite{Perros2017-dh}
             \State $\rho=trace(\M{G})/R$
             \State $\M{L}=$Cholesky$(\M{G}+\rho \M{I})$
             \While{convergence criterion is not met}
                  \State $\M{Z_n}^T=(\M{L^T})^{-1}\M{L}^{-1}(\M{F}+\rho(\overbar{\M{Z_n}}+\M{D}_{Z_n}))^T$
                  \State $\overbar{\M{Z_n}}:=\arg\min_{\overbar{\M{Z_n}}} c(\overbar{\M{Z_n}}) +\frac{\rho}{2} ||\overbar{\M{\M{Z_n}}}-\M{Z_n^T}+\M{D}_{Z_n}||_F^2 $
                  \State $\M{D}_{Z_n}:=\M{D}_{Z_n}+\overbar{\M{Z_n}}-\M{Z_n}^T$
          \EndWhile

    \EndFor
    
 \EndWhile
 \State $\M{H}=\overbar{\M{Z_1}}$, $W=\overbar{\M{Z_2}}$, $\M{V}=\overbar{\M{Z_3}}$
 \For{k=1,...,K }
          
          \State $\M{U_k}=\M{Q_k} \M{H}$   
          \State $\M{S_k}=diag(\V{W(k,:)})$   

    \EndFor
\end{algorithmic}
\label{CONSPAR2_alg}
\end{algorithm}

\subsection{Examples of useful constraints} \label{sec:constraints}
Next, we describe several special constraints which are useful for many applications and derive the updates rules for those constraints.
\subsubsection{\textbf{Smoothness on {$U_k$}}:} 
For longitudinal data such as EHRs, imposing latent components that change smoothly over time may be desirable to improve interpretability and robustness (less fitting to noise). Motivated by previous work \cite{helwig2017estimating,timmerman2002three}, we incorporate temporal smoothness to the factor matrices {\small$\M{U_k}$} by approximating them as a linear combination of several smooth functions.
In particular, we use M-spline, a non-negative spline function
which can be efficiently computed through a recursive formula \cite{ramsay1988monotone}.
For each subject $k$, a set of M-spline basis functions  ({\small$\M{M_k} \in \mathbb{R}^{I_k \times l} $}) are created where {\small$l$} is the number of basis functions.
Thus, {\small$\M{U_k}$} is an unknown linear combination of the smooth basis functions {\small$\M{U_k} = \M{M_k} \M{W_k}$}, where {\small$\M{W_k}$} is the unknown weight matrix.

The temporal smoothness constrained solution is equivalent to performing the PARAFAC2 algorithm on a projected {\small$\M{X_k^{\textprime}}=\M{C_k^T}\M{X_k}$}, where {\small$\M{C_k}$ }is obtained from the SVD of {\small$\M{M_k} = [\M{C_k},\M{O_k},\M{P_k^T}]$}.
We provide proof of the equivalence by analyzing the update of {\small$\M{Q_k}$} for the newly projected data ({\small$\M{X_k^{\textprime}}$}): 
\begin{equation}
\small
\begin{aligned}
& \underset{\M{Q_k}}{\text{minimize}}
& & ||\M{C_k^T}\M{X_k} -\M{Q_k}\M{H}\M{S_k}\M{V^T}||_F^2
\end{aligned}
\end{equation}
This can be re-written as the minimization of {\small$tr(\M{X_k^T}\M{C_k}\M{C_k^T}\M{X_k})-2tr(\M{X_k^T}\M{C_k}\M{Q_k}\M{H}\M{S_k}\M{V^T})+\\tr(\M{V}\M{S_k}\M{H^T}
\M{{Q_k}^T}\M{Q_k}\M{H}\M{S_k}\M{V^T})$}. Since {\small$\M{C_k}$} and {\small $\M{Q_k}$} have orthonormal columns  the first and third terms are constants. Also  $tr(\M{A^T})=tr(\M{A})$ and $tr(\M{A}\M{B}\M{C})=tr(\M{C}\M{A}\M{B})=tr(\M{B}\M{C}\M{A})$, thus the update is equivalent to: 
\begin{equation}
\small
\begin{aligned}
& \underset{\M{Q_k}}{\text{max}}
& & tr(\M{C_k^T}\M{X_k}\M{V}\M{S_k}\M{H^T}\M{{Q_k}^T})=tr(\M{X_k^{\textprime}}\M{V}\M{S_k}\M{H^T}\M{{Q_k}^T}) \\
\iff	& \underset{\M{Q_k}}{\text{min}}
& & ||\M{X_k^{\textprime}}\M{V}\M{S_k}\M{H^T} -\M{Q_k}||_F^2
\end{aligned}
\end{equation}
This is similar to equation \ref{solv_Q} (only difference is {\small$\M{X_k^{\textprime}}$}, a projection of {\small$\M{X_k}$}) which can be solved using the constrained quadratic problem \cite{golub2013matrix}.
Thus, solving for {\small$\M{H}$}, {\small$\M{S_k}$}, and {\small$\M{V}$} remains the same. The only difference is that after convergence, {\small$\M{U_k}$} is constructed as {\small$\M{C_k}\M{Q_k}\M{H}$}. 

\begin{figure}
\centering

\begin{subfigure}[b]{0.53\textwidth}
\centering
\begin{tikzpicture}[scale=.4]
    \begin{axis}[
    width=19cm,
    height=7cm,
    legend style={font=\fontsize{12}{5}\selectfont},
    xlabel={Hospital Visits},
    ylabel={Magnitude},
    label style={font=\Large},
    xmin=1, xmax=81,
    ymin=0, ymax=0.2,
    legend pos=outer north east,
    legend columns=4,
    legend style={at={(0.5,1.3)},anchor=north},
    xmajorgrids=true,
    grid style=dashed,
    ]
    \addplot  [color=red,very thick] table[x=visit,y=M1] {results/M_spline_gap.txt}; \addlegendentry{Basis function 1}
     \addplot  [color=blue,very thick] table[x=visit,y=M2] {results/M_spline_gap.txt}; \addlegendentry{Basis function 2}
            
                \addplot  [color=green,very thick ] table[x=visit,y=M3] {results/M_spline_gap.txt}; \addlegendentry{Basis function 3}
                    \addplot  [color=yellow,very thick] table[x=visit,y=M4] {results/M_spline_gap.txt}; \addlegendentry{Basis function 4}
                      \addplot  [color=black,very thick] table[x=visit,y=M5] {results/M_spline_gap.txt}; \addlegendentry{Basis function 5}
                        \addplot  [color=magenta,very thick] table[x=visit,y=M6] {results/M_spline_gap.txt}; \addlegendentry{Basis function 6}
                            \addplot  [color=cyan,very thick] table[x=visit,y=M7] {results/M_spline_gap.txt}; \addlegendentry{Basis function 7}
    \end{axis}
\end{tikzpicture}
\caption{\centering Basis functions used by \methodName.}
\label{mspline_no_gap}
\end{subfigure}%
\begin{subfigure}[b]{0.53\textwidth}
\begin{tikzpicture}[scale=.4]
    \begin{axis}[
    width=19cm,
    height=7cm,
    legend style={font=\fontsize{12}{5}\selectfont},
    xlabel={Hospital Visits},
    ylabel={Magnitude},
    label style={font=\Large},
    xmin=1, xmax=81,
    ymin=0, ymax=0.2,
    legend pos=south east,
    xmajorgrids=true,
    grid style=dashed,
    ]
    \addplot  [color=red,very thick] table[x=visit,y=M1] {results/M_spline_no_gap.txt}; 
     \addplot  [color=blue,very thick] table[x=visit,y=M2] {results/M_spline_no_gap.txt}; 
            
                \addplot  [color=green,very thick] table[x=visit,y=M3] {results/M_spline_no_gap.txt}; 
                    \addplot  [color=yellow,very thick] table[x=visit,y=M4] {results/M_spline_no_gap.txt}; 
                      \addplot  [color=black,very thick] table[x=visit,y=M5] {results/M_spline_no_gap.txt}; 
                        \addplot  [color=magenta,very thick] table[x=visit,y=M6] {results/M_spline_no_gap.txt}; 
                            \addplot  [color=cyan,very thick] table[x=visit,y=M7] {results/M_spline_no_gap.txt};

    \end{axis}
\end{tikzpicture}
\caption{\centering Basis functions used by Helwig.}
\label{mspline_gap}
\end{subfigure}
 \caption{\footnotesize 7 Basis functions for a patient with sickle cell anemia. Figure \ref{mspline_no_gap} shows the basis functions that \methodName used for incorporating the smoothness that considers the gap between two visits while figure \ref{mspline_gap} related to basis functions for Helwig which divide the range [0,80] based on a equal distance.}
\label{fig:M_bspline}
\end{figure}
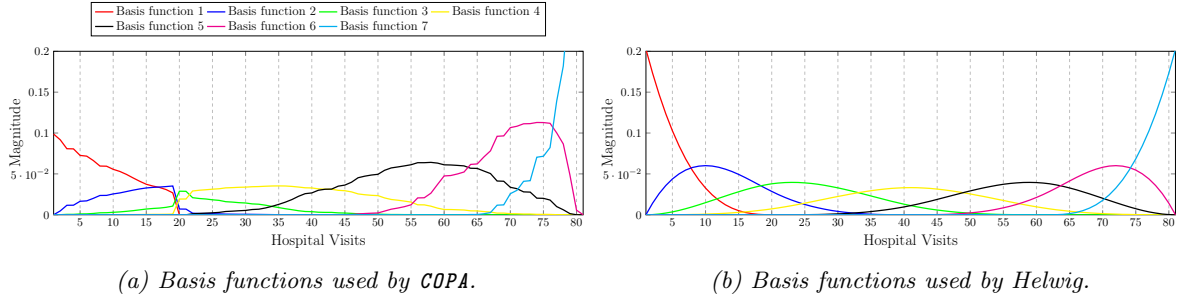

In some domains, there may be uneven time gaps between the observations.
For example, in our motivating application, patients may not regularly visit a healthcare provider but when they do, the visits are closely clustered together.
To adaptively handle time-varying gaps, we alter the basis functions to account for uneven gaps.
Under the assumption of small and equidistant gaps (Helwig's approach), the basis functions are created directly on the visits.
Instead, we define a set of M-spline functions for each patient in the interval [$t_1$, $t_n$], where {\small$t_1$} and {\small$t_n$} are the first and last hospital visits.
These spline functions (with day-resolution) are then transformed to their visit-resolution.
Thus, given the number and position of the knots ({\small$[\beta_1. .\beta_m]$}), which can be estimated using the metric introduced in Helwig's work \cite{helwig2017estimating}, we create the {\small$i^{th}$} basis function of patient {\small$k$} with degree {\small$d$} using the following recursive formula:
\[\small
m_{ik,d}(t)=\frac{t-\beta_i}{\beta_{i+d}-\beta_i}m_{ik,d-1}(t)+\frac{\beta_{i+d+1}-t}{\beta_{i+d+1}-\beta_{i+1}}m_{i+1k,d-1}(t)
\]
where $t$ denotes the hospital visit day and {\small$\beta_i$} is the {\small$i^{th}$} knot.
Hence, we can reconstruct the basis functions as {\small$m_{ik,0}(t)$} is 1 if {\small$t \in [\beta_i,\beta_{i+1}]$} and zero otherwise.  Figure \ref{fig:M_bspline} shows the two types of basis functions related to a patient with sickle cell anemia.
\pgfplotstableread[col sep=space]{results/M_spline_gap.txt}\datatable
\pgfplotstableread[col sep=space]{results/M_spline_no_gap.txt}\datatable

\subsubsection{\textbf{Sparsity on {$V$}:}}
Sparsity constraints have wide applicability to many different domains and have been exploited for several purposes including improved interpretability, reduced model complexity, and increased robustness.
For the purpose of EHR-phenotyping, we impose sparsity on factor matrix {\small$\M{V}$}, to obtain sparse phenotype definitions.
While several sparsity inducing constraints can be introduced, we focus on the {\small$\ell_0$} and {\small$\ell_1$} norms, two popular regularization techniques. The {\small$\ell_0$} regularization norm, also known as hard thresholding, is a non-convex optimization problem that caps the number of non-zero values in a matrix. 
 The {\small$\ell_1$} regularization norm, or the soft thresholding metric, is often used as a convex relaxation of the {\small$\ell_0$} norm.
The objective function with respect to {\small$\M{V}$} for the sparsity ({\small$\ell_0$} norm) constrained PARAFAC2 is as follows:
\begin{equation}
\small
\begin{aligned}
& \underset{\M{V}}{\text{minimize}}
& &  \frac{1}{2}||\T{Y}-[\M{H};\M{V};\M{W}]||_F^2 +  \lambda ||\overbar{\M{V}}||_0 \text{,~s.t.~} \M{V}=\overbar{\M{V}}
\end{aligned}
\label{Solve_Sparse_V}
\end{equation}
where {\small$\lambda$} is a regularization parameter which needs to be tuned. The proximity operator for the {\small$\ell_0$} regularization, {\small$\overbar{\M{V}}$}, uses the hard-thresholding operator which zeros out entries below a specific value. Thus, the update rules for {\small$\M{V}$} are as follows:
\begin{equation*}
\small
\begin{aligned}
\M{V^T}&:=\Big((\M{H^T} \M{H} * \M{W^T} \M{W})+ \rho \M{I}\Big)^{-1}(\M{Y_{(2)}}(\M{W}\odot \M{H})+\rho(\overbar{\M{V}}+\M{D}_{V^T}))^T \\
\overbar{\M{V}}:&= \arg\min_{\overbar{\M{V}}} \lambda||\overbar{\M{V}}||_0+\frac{\rho}{2} ||\overbar{\M{V}}-\M{V^T}-\M{D}_{V^T}||_F^2= \left\{
    \begin{array}{ll}
       0 & \overbar{\M{V}}^2 \leq \frac{2\lambda}{\rho}=\mu\\
        \overbar{\M{V}} & \overbar{\M{V}}^2 \geq \frac{2\lambda}{\rho}=\mu
    \end{array}
\right. 
\\
\M{D}_{V^T} &:=\M{D}_{V^T}+\overbar{\M{V}}-\M{V^T}
\end{aligned}
\end{equation*}
The update rule corresponding to the {\small$\ell_1$} regularization, {\small$\overbar{\M{V}}$}, is the soft-thresholding operator: 
\begin{equation*}
\small
\begin{aligned}
\overbar{\M{V}}:&= \arg\min_{\overbar{\M{V}}}  \lambda||\overbar{\M{V}}||_1+\frac{\rho}{2} ||\overbar{\M{V}}-\M{V^T}-\M{D}_{V^T}||_F^2 = \max( 0, |\M{V^T}+\M{D}_{V^T}| - (\lambda/\rho) )
\end{aligned}
\end{equation*}

Note that imposing factor sparsity boils down to using element-wise thresholding operations, as can be observed above. Thus, imposing sparsity is scalable even for large datasets.

\subsubsection{\textbf{Non-negativity on { $S_k$}}:}
\methodName~ is able to impose non-negativity constraint to factor matrices {\small$\M{H}$}, {\small$\M{S_k}$}, and {\small$\M{V}$}. Because the updating rules for these three factor matrices are same we just show the update rules for factor matrix {\small$\M{S_k}$}  for simplicity ({\small$\M{S_k}=diag(\V{W(k,:)})$}):
\begin{equation*}
\small
\begin{aligned}
\M{W^T} &:=\Big((\M{H^T} \M{H} * \M{V^T} \M{V})+ \rho \M{I}\Big)^{-1}(\M{Y_{(3)}}(\M{V}\odot \M{H})+\rho(\overbar{\M{W}}+\M{D}_{W^T}))^T \\
\overbar{\M{W}} &:=max(0, \M{W^T}-\M{D}_{W^T} ) \\
\M{D}_{W^T} &:=\M{D}_{W^T}+\overbar{\M{W}}-\M{W^T}
\end{aligned}
\end{equation*}
Note that our update rule for {\small$\overbar{\M{W}}$} only involves zeroing out the negative values and is an element-wise operation. The alternating least squares framework proposed by \cite{bro1999parafac2} and employed by SPARTan \cite{Perros2017-dh} can also achieve non-negativity through non-negative least squares algorithms but that is a more expensive operation than our scheme. 

\section{Experimental Results}
In this section, we first provide the description of the real datasets. Then we give an overview of baseline methods and evaluation metrics. After that, we present the quantitative experiments. 
Finally, we show the success of our algorithm in discovering temporal signature of patients and phenotypes on a subset of medically complex patients from a real data set.
\subsection{Setup}
\subsubsection{\textbf{Data Set Description}}
\par \textbf{Children's Healthcare of Atlanta
(CHOA): }
This dataset contains the EHRs of 247,885 pediatric patients with at least 3 hospital visits. For each patient, we utilize the International Classification of Diseases (ICD9) codes \cite{slee1978international}  and medication categories from their records, as well as the provided age of the patient (in days) at the visit time. To improve interpretability and clinical meaningfulness, ICD9 codes are mapped into broader Clinical Classification Software (CCS) \cite{ccs}  categories.
Each patient slice {\small$\M{X}_k$} records the clinical observations and the medical features.
The resulting tensor is 247,885 patients by 1388 features by maximum 857 observations.


\textbf{Centers for Medicare and Medicaid (CMS):}\footnote{\url{https://www.cms.gov/Research-Statistics-Data-and-Systems/Downloadable-Public-Use-Files/SynPUFs/DE_Syn_PUF.html}} CMS released the Data Entrepreneurs Synthetic Public Use File (DE-SynPUF), a realistic set of claims data that also protects the Medicare beneficiaries' protected health information. The dataset is based on 5\% of the Medicare beneficiaries during the period between 2008 and 2010. 
Similar to CHOA, we extracted ICD9 diagnosis codes and summarized them into CCS categories. The resulting number of patients are 843,162 with 284 features and the maximum number of observations for a patient are 1500.

Table \ref{tab:summry_data} provides the summary statistics of real  datasets. 

\begin{table}[htbp]
  \centering
  \caption{\footnotesize Summary statistics of real  datasets that we used in the experiments. $K$ denotes the number of patients, $J$ is the number of medical features and $I_k$ denotes the number of clinical visits for $k^{th}$ patient. }
   \scalebox{0.80}{
    \begin{tabular}{lllll}
    \multicolumn{1}{l}{Dataset} & \multicolumn{1}{l}{$K$} & \multicolumn{1}{l}{$J$} & \multicolumn{1}{l}{max($I_k$)} & \multicolumn{1}{l}{\#non-zero elements} \\
    \midrule
    CHOA  & 247.885 & 1388  & 857   & 11 Million \\
    CMS  & 843,162 & 284   & 1500  & 84 Million \\
    \end{tabular}%
    }
  \label{tab:summry_data}%
\end{table}%

\subsubsection{\textbf{Baseline Approaches}}
In this section, we briefly introduce the baseline that we compare our proposed method. 
\begin{itemize}
\item \textbf{SPARTan \cite{Perros2017-dh}}\footnote{The MATLAB code is  available at \url{https://github.com/kperros/SPARTan}} is a recently-proposed methodology for fitting PARAFAC2 on large and sparse data. The algorithm reduces the execution time and memory footprint of the bottleneck MTTKRP operation. Each step of SPARTan updates the model factors in the same way as the classic PARAFAC2 model \cite{kiers1999parafac2}, but is faster and more memory efficient for large and sparse data. In the experiments, SPARTan has non-negativity constraints on  $\M{H}$, $\M{S_k}$, and $\M{V}$ factor matrices.
\item \textbf{Helwig \cite{helwig2017estimating}}
incorporates smoothness into PARAFAC2 model by constructing a library of smooth  functions for every subject and apply smoothness based on linear combination of library functions. We implemented this algorithm in MATLAB.
\end{itemize}

\subsubsection{\textbf{Evaluation Metrics}}
We use  FIT \cite{bro1999parafac2} to evaluate the quality of the reconstruction based on the model's factors: 
\[
\small
FIT= 1-\frac{\sum_{k=1}^{K} ||\M{X_k} -\M{U_k} \M{S_k}\M{V^T}||^2}{\sum_{k=1}^{K} ||\M{X_k}||^2}
\]
The range of FIT is between {\small$[-\infty,1]$} and values near 1 indicate the method can capture the data perfectly.
We also use SPARSITY metric to evaluate the factor matrix {\small$\M{V}$} which is as follows:
\[
\small
SPARSITY= \frac{nz(\M{V})}{size(\M{V})}
\]
where $nz(V)$ is the number of zero elements in {\small$\M{V}$} and $size(\M{V})$ is the number of elements in {\small$\M{V}$}. Values near 1 implies the sparsest solution.

\subsubsection{\textbf{Implementation details}}
\methodName~ is implemented in MATLAB and includes functionalities from the Tensor Toolbox \cite{TTB_Software}. To enable reproducibility and broaden the usage of the PARAFAC2 model, our implementation is \textit{publicly available at: }{\url{https://github.com/aafshar/COPA}}. All the approaches (including the baselines) are evaluated on MatlabR2017b. We also implemented the smooth and functional PARAFAC2 model \cite{helwig2017estimating}, as the original approach was only available in R \cite{helwig2017package}. This ensures a fair comparison with our algorithm.

\subsubsection{\textbf{Hardware}}
The experiments were all conducted on a server running Ubuntu 14.04 with 250 GB of RAM and four Intel E5-4620 v4 CPU's with a maximum clock frequency of 2.10GHz. Each processor contains 10 cores. Each core can exploit 2 threads with hyper-threading enabled.

\subsubsection{\textbf{Parallelism}}
We utilize the capabilities of Parallel Computing Toolbox of Matlab by activating parallel pool for all methods. 
For CHOA  dataset, we used 20 workers whereas for CMS  we used 30 workers because of  more number of non-zero values. 

\subsection{Quantitative Assessment of Constraints}
\begin{figure}[b]
\centering
\includegraphics[width=.55\textwidth]{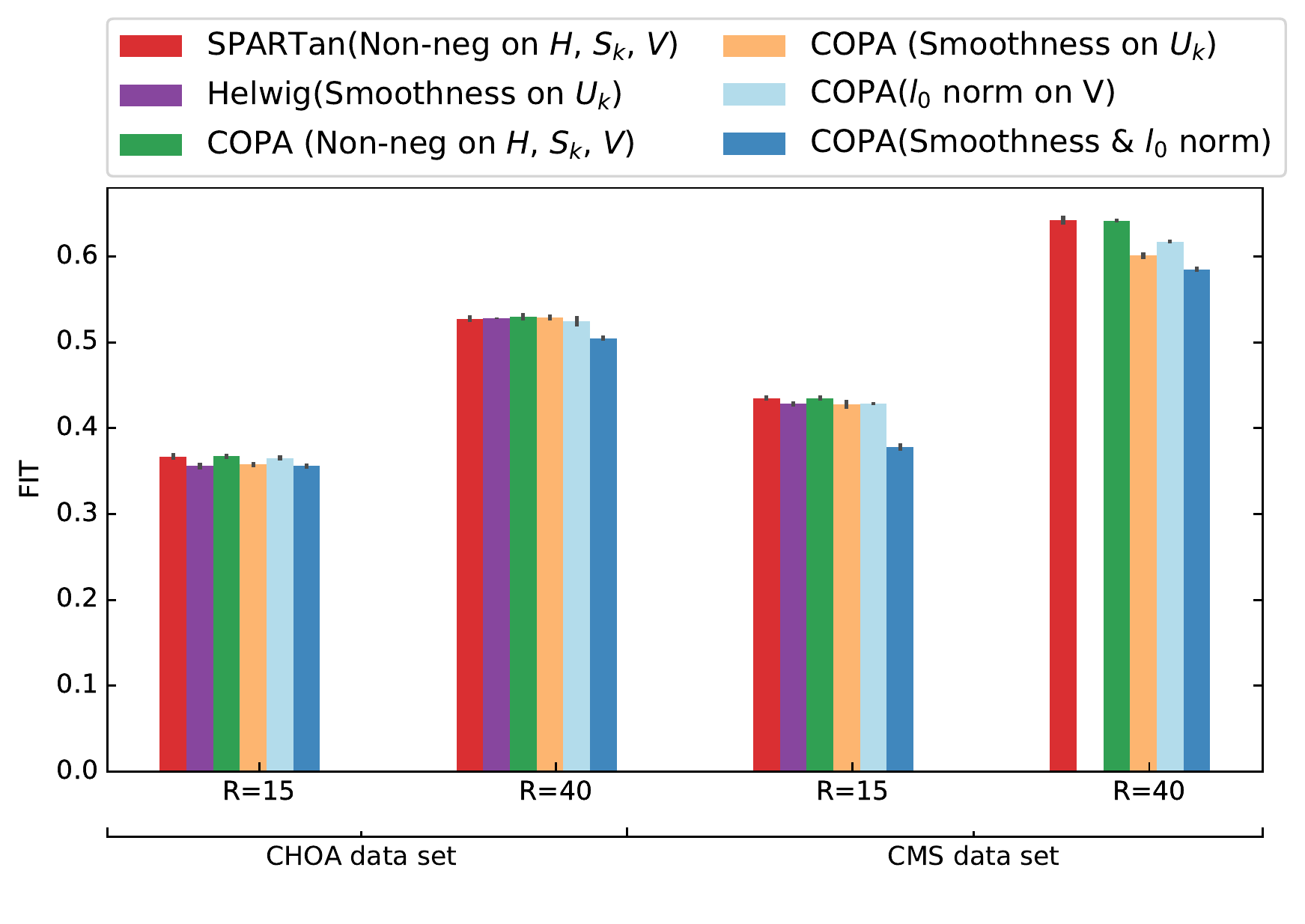}
\caption{\footnotesize Comparison of FIT for different approaches with various constraints on two target ranks $R=15$ and $R=40$ on real world datasets. Overall, \mname achieves comparable fit to SPARTan while supporting more constraints. The missing purple bar in the forth column is out of memory failure for Helwig method.}
\label{fig:FIT_diff_ALG}
\end{figure}

 To understand how  different constraints affect the reconstruction error, we perform an experiment using each of the constraints introduced in Section \ref{sec:constraints}. We run each method for 5 different random initializations and provide the average and standard deviation of FIT as shown in Figure \ref{fig:FIT_diff_ALG}. 
This Figure  illustrates the impact of each constraint on the FIT values across both datasets for two different target ranks (R=\{15,40\}). In all versions of \methodName, $\M{S_k}$ factor matrix is non-negative. Also, we apply smoothness on {\small$\M{U_k}$} and {\small$\ell_0$} regularization norm on {\small$\M{V}$} separately and also simultaneously. From Figure \ref{fig:FIT_diff_ALG}, we observe that  different versions of \methodName can produce a comparable value of FIT even with both smoothness on {\small$\M{U_k}$} and sparsity on {\small$\M{V}$}.
The number of smooth basis functions are selected based on the cross-validation metric introduced in \cite{timmerman2002three}  and the {\small$\ell_0$} regularization parameter ({\small$\mu$}) is selected via grid search by finding a good trade off between FIT and SPARSITY metric. The optimal values of each parameter for the two different data sets and target ranks are reported in table \ref{table:param}.

\begin{table}[h]
    \centering
    \caption{Values of parameters ($l$, $\mu$) for different data sets and various target ranks for \methodName.}
    \label{table:param}
    
  \scalebox{0.7}{
    \begin{tabular}{ l c c c c c }
    \toprule
    &  \multicolumn{2}{c}{CHOA} & \multicolumn{2}{c}{CMS}  \\
    Algorithm & R=15 & R=40 & R=15 & R=40  \\ 
    \midrule
    \# basis functions ($l$) & 33& 81    & 106 & 253  \\ 
    $\mu$   & $23$ &  $25$   & $8$& $9$ \\
    \bottomrule
    \end{tabular}
    }
\end{table}

We next quantitatively evaluate the effects of  sparsity (average and standard deviation of the sparsity metric) by applying {\small$\ell_0$} regularization norm  on the factor matrix $\M{V}$ for \methodName~ and compare it with SPARTan for 5 different random initializations, as provided in Table \ref{table:SparsityCompare}.
For both the CHOA and CMS datasets, \methodName~achieves more than a 98\% sparsity level.
The improved sparsity of the resulting factors is especially prominent in the CMS dataset, with a 400\% improvement over SPARTan.
Sparsity can improve the interpretability and potentially the clinical meaningfulness of phenotypes via more succinct patient characterizations.
The quantitative effectiveness is further supported by the qualitative endorsement of a clinical expert (see Section \ref{sec:pheno_disc}).

\begin{table}[h]
    \centering
    \caption{
    The average and standard deviation of sparsity metric (fraction of zero elements divided by the matrix size) comparison for the factor matrix $V$ on CHOA and CMS using two different target ranks for 5 different random initializations.}
    \label{table:SparsityCompare}
    
  \scalebox{0.75}{
    \begin{tabular}{ l c c c c c }
    \toprule
    &  \multicolumn{2}{c}{CHOA} & \multicolumn{2}{c}{CMS}  \\
    Algorithm & R=15 & R=40 & R=15 & R=40  \\ 
    \midrule
    \methodName~ & 0.9886\textpm 0.0035& 0.9897\textpm 0.0027   &0.9950\textpm 0.0001 & 0.9963\textpm 0.0002  \\ 
    SPARTan \cite{Perros2017-dh}   & 0.7127 \textpm 0.0161 &  0.8127\textpm     0.0029
   & 0.1028\textpm 0.0032& 0.2164 \textpm 0.0236 \\
    \bottomrule
    \end{tabular}
    }
\end{table}


\begin{figure}[ht]
\centering
    \begin{subfigure}{1.0\linewidth}
  	\centering\includegraphics[width=.7\linewidth]{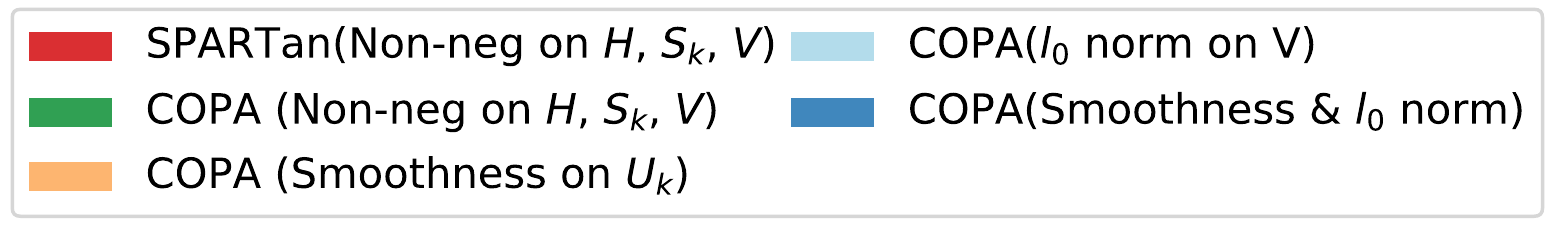}
  \end{subfigure}\\
    \begin{subfigure}[t]{0.3\textwidth}
        \includegraphics[height=1.5in]{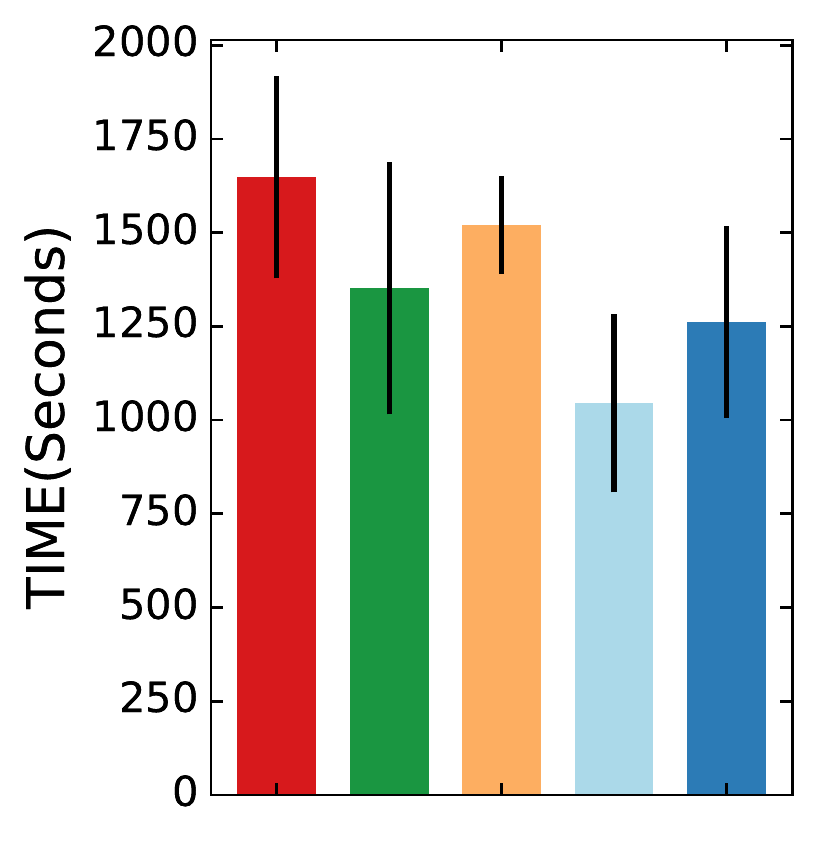}
        \caption{CHOA, R=15}
    \end{subfigure}%
    \begin{subfigure}[t]{0.3\textwidth}
        \includegraphics[height=1.5in]{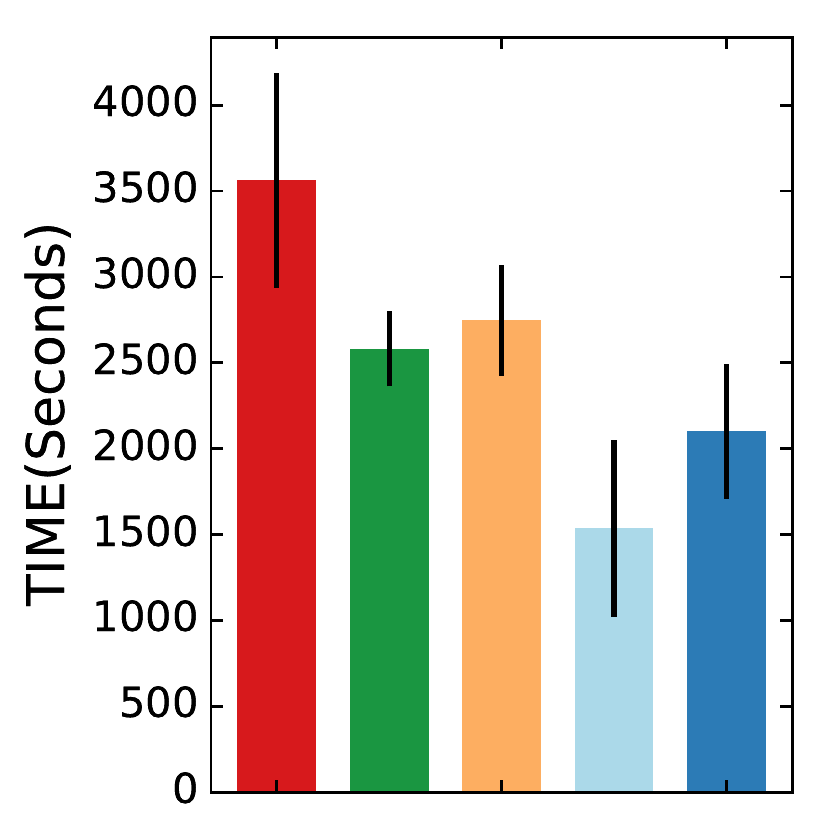}
        \caption{CHOA, R=40}
    \end{subfigure}\\
    \begin{subfigure}[t]{0.3\textwidth}
        \includegraphics[height=1.5in]{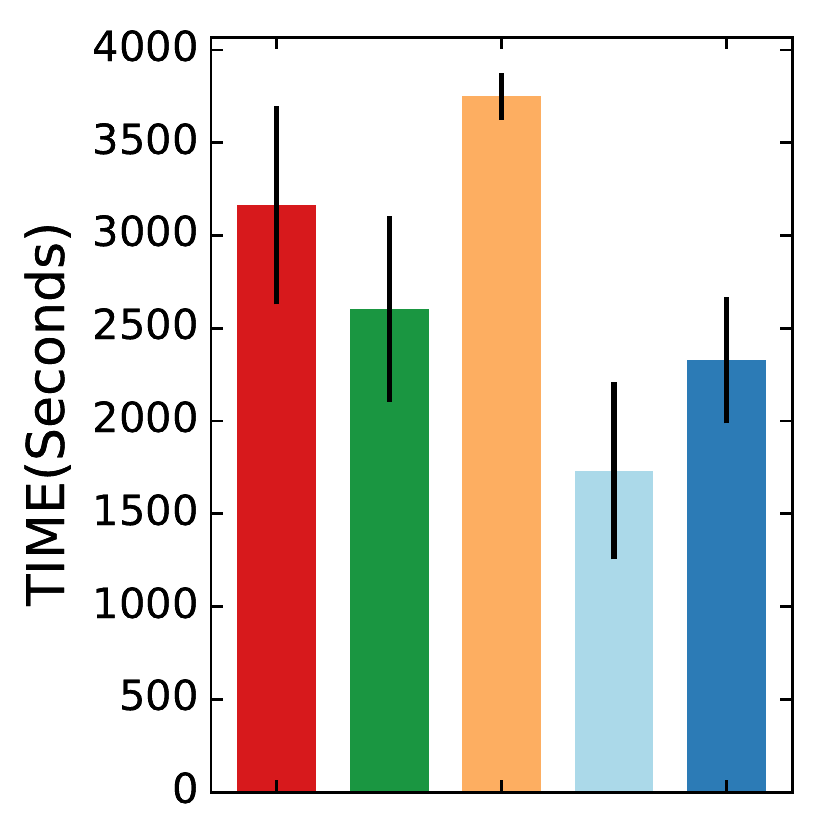}
        \caption{CMS, R=15}
    \end{subfigure}%
    \begin{subfigure}[t]{0.3\textwidth}
        \includegraphics[height=1.5in]{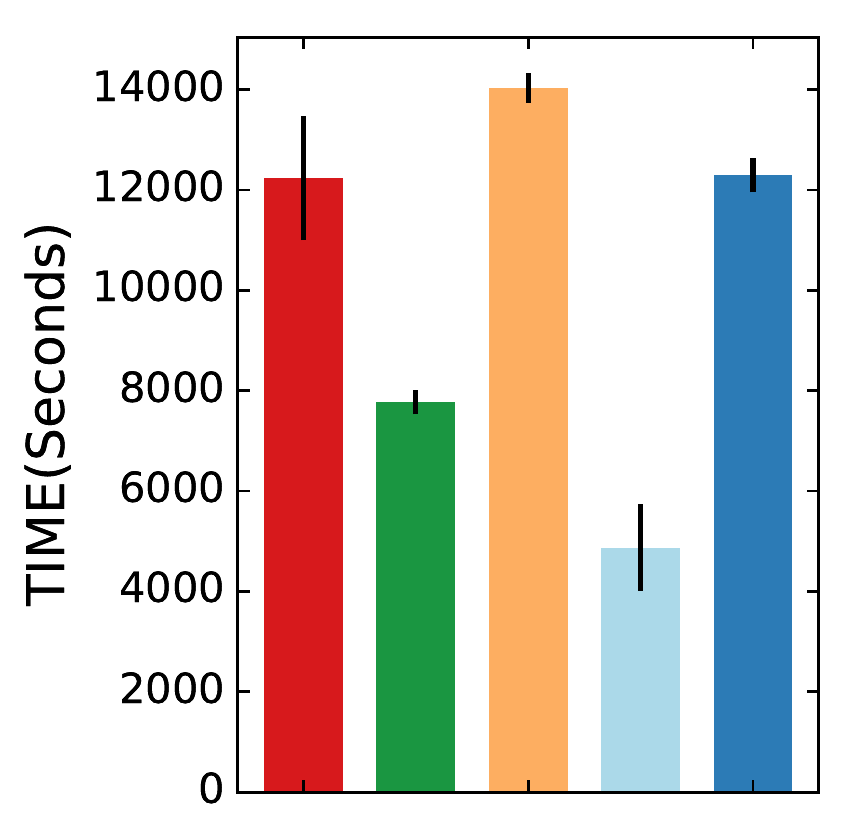}
        \caption{CMS, R=40}
    \end{subfigure}
    \caption{\footnotesize The Total Running Time comparison (average and standard deviation) in seconds for different versions of \methodName and SPARTan for 5 different random initializations. Note that even with smooth constraint \mname performs just slightly slower than SPARTan, which does not support such smooth constraints.  }
    \label{fig:time_comp}
\end{figure}

\subsection{Scalability and FIT-TIME}
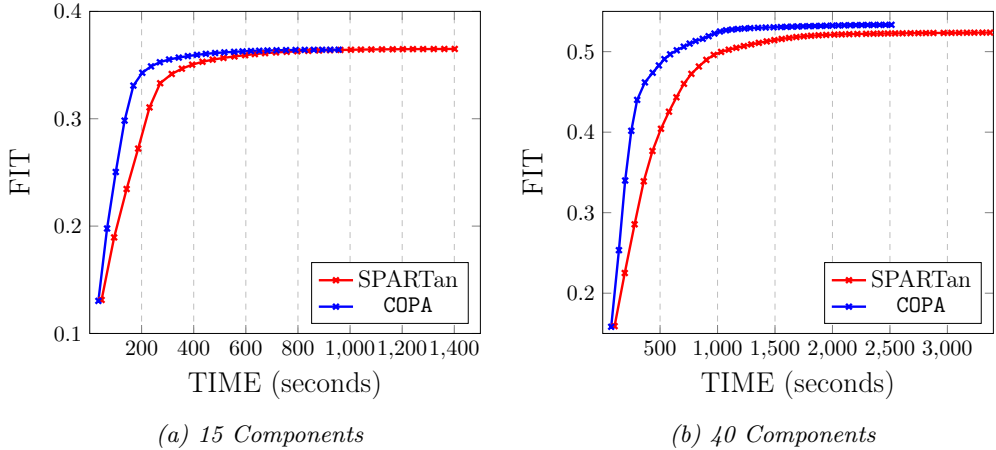
\begin{figure}[!t]
\centering

\begin{subfigure}[b]{0.45\textwidth}

\begin{tikzpicture}[scale=.75]
    \begin{axis}[
    legend style={font=\fontsize{12}{5}\selectfont},
    xlabel={TIME (seconds)},
    ylabel={FIT},
    label style={font=\Large},
    xmin=1, xmax=1500,
    ymin=0.1, ymax=0.4,
    legend pos=south east,
    xmajorgrids=true,
    grid style=dashed,
    ]
    \addplot  [color=red,very thick, mark=x,] table[x=TIME,y=Relative_Error] {CHOA/SPARTan_15.txt}; \addlegendentry{SPARTan}
    
    \addplot  [color=blue,very thick, mark=x,] table[x=TIME,y=Relative_Error] {CHOA/AOADMM_15.txt}; \addlegendentry{\methodName~}

    \end{axis}
\end{tikzpicture}
\caption{15 Components}
\label{fig:5comp}
\end{subfigure}%
\begin{subfigure}[b]{0.45\textwidth}
\begin{tikzpicture}[scale=.75]
    \begin{axis}[
    legend style={font=\fontsize{12}{5}\selectfont},
    xlabel={TIME (seconds)},
    ylabel={FIT},
    label style={font=\Large},
    xmin=1, xmax=3400,
    ymin=0.15, ymax=0.55,
    legend pos=south east,
    xmajorgrids=true,
    grid style=dashed,
    ]
    \addplot  [color=red,very thick, mark=x,] table[x=TIME,y=Relative_Error] {CHOA/SPARTan_40.txt}; \addlegendentry{SPARTan}
    
    \addplot  [color=blue,very thick, mark=x,] table[x=TIME,y=Relative_Error] {CHOA/AOADMM_40.txt}; \addlegendentry{\methodName~}

    \end{axis}
\end{tikzpicture}

\caption{40 Components}
\label{fig:temp-apr}
  \end{subfigure}

\caption{\footnotesize The best Convergence of \methodName~ and SPARTan out of 5 different random initializations with non-negativity constraint on $H$, $\{S_k\}$, and $V$  on CHOA data set  for different target ranks (two cases considered: R=\{15,40\}).}
\label{fig:conv_D1}
\end{figure}

\begin{figure}
\centering

\begin{subfigure}[b]{0.4\textwidth}
\begin{tikzpicture}[scale=.65]
    \begin{axis}[
    legend style={font=\fontsize{12}{5}\selectfont},
    xlabel={TIME (seconds)},
    ylabel={FIT},
    label style={font=\Large},
    xmin=1, xmax=3000,
    ymin=0.18, ymax=0.446,
    legend pos=south east,
    xmajorgrids=true,
    grid style=dashed,
    ]
    \addplot  [color=red,very thick, mark=x,] table[x=TIME,y=Relative_Error] {CMS/SPARTan_15.txt}; \addlegendentry{SPARTan}
    
     \addplot  [color=blue,very thick, mark=x,] table[x=TIME,y=Relative_Error] {CMS/AOADMM_15.txt}; \addlegendentry{\methodName}

    \end{axis}
\end{tikzpicture}
\caption{  15 Components.}
\end{subfigure}%
\begin{subfigure}[b]{0.4\textwidth}
\begin{tikzpicture}[scale=.65]
    \begin{axis}[
    legend style={font=\fontsize{12}{5}\selectfont},
    xlabel={TIME (seconds)},
    ylabel={FIT},
    label style={font=\Large},
    xmin=1, xmax=11800,
    ymin=0.25, ymax=0.654,
    legend pos=south east,
    xmajorgrids=true,
    grid style=dashed,
    ]
    \addplot  [color=red,very thick, mark=x,] table[x=TIME,y=Relative_Error] {CMS/SPARTan_40.txt}; \addlegendentry{SPARTan}
    
     \addplot  [color=blue,very thick, mark=x,] table[x=TIME,y=Relative_Error] {CMS/AOADMM_40.txt}; \addlegendentry{\methodName}

    \end{axis}
\end{tikzpicture}
\caption{ 40 Components.}
\label{fig:50_S01_synth}
\end{subfigure}
 \caption{\footnotesize The best convergence of \methodName~ and SPARTan out of 5 different random initializations with non-negativity constraint on $H$, $\{S_k\}$, and $V$ on CMS data with K=843,162, J=284 and maximum number of observations are 1500. Algorithms tested on different target ranks (two cases considered: R=\{15,40\}).}
\label{fig:Conv_D2}
\end{figure}
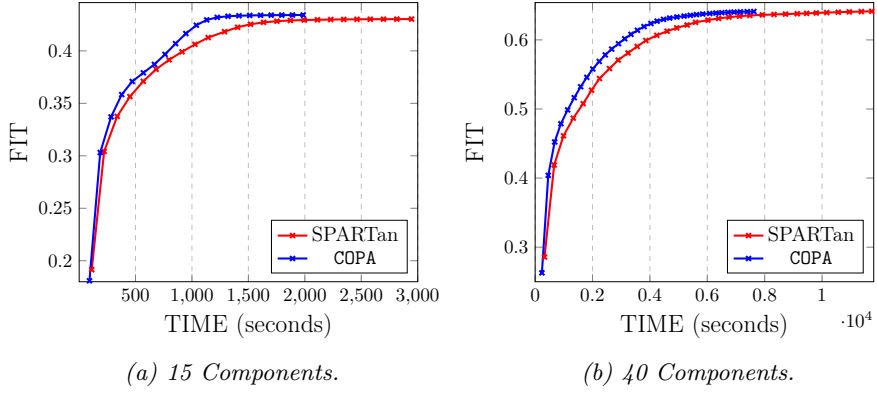

First, we evaluate and compare the total running time of all versions of COPA and SPARTan on the real datasets. We run each method 5 times and report averages and standard deviations. As shown in Figure \ref{fig:time_comp}, the average of total running time of \methodName with non-negativity constraints imposed on $\M{H}, \{\M{S_k}\}, \M{V}$ is faster (up to 1.57$\times$) than SPARTan with the same set of constraints for two data sets and different target ranks. In order to provide more precise comparison we apply paired t-tests on the two sets of running time, one from SPARTan and the other from a version of \methodName under the null hypothesis that the running times are not significantly different between the two methods. We present the p-values return from the t-tests in Table \ref{tab:t-test}. The p-values for COPA with non-negativity constraint and sparsity constraint are  small which suggest that the version of \methodName is significantly better than the SPARTan (rejecting the null hypothesis). Also we provide the speedups (running time of SPARTan divide by running time of \methodName) in Table \ref{tab:t-test}. 
 Moreover, the average running times of Smooth \methodName  are just slightly slower than SPARTan, which does not support such smooth constraints. 
 Next, we compare the best convergence (Time in seconds versus FIT) out of 5 different random initializations  of the proposed \methodName~approach against SPARTan. 
For both methods, we add non-negativity constraints to {\small$\M{H}$}, {\small$\{\M{S_k}\}$}, {\small$\M{V}$} and compare the convergence rates on both real-world datasets for two different target ranks ({\small$R=\{15,40\}$}).
Figures \ref{fig:conv_D1} and \ref{fig:Conv_D2} illustrates the results on the CHOA and CMS datasets respectively.
\methodName~ converges faster than SPARTan in all cases.
While both \methodName~ and SPARTan avoid direct construction of the sparse tensor {\small$\T{Y}$}, the computational gains can be attributed to the efficiency of the non-negative proximity operator, an element-wise operation that zeros out the negative values in \methodName whereas SPARTan performs expensive NN-Least Square operation. Moreover,  caching the MTTKRP operation  and the Cholesky decomposition of the Gram matrix help \methodName to reduce the number of computations.


\begin{table}[htbp]
  \centering
  \caption{Speedups (running time of SPARTan divided by running time of \methodName for various constraint configurations) and corresponding p-values. \methodName is faster (up to 2.5$\times$) on the majority of constraint configurations as compared to the baseline SPARTan approach which can only handle non-negativity constraints.}
  \scalebox{0.65}{

    \begin{tabular}{c|c|cccc}
    \multicolumn{1}{c}{} & \multicolumn{1}{c}{} & Non-neg COPA & Smooth COPA & Sparse COPA & Smooth \& Sparse COPA \\
    \midrule
    \multirow{2}[2]{*}{CHOA, R=15} & Speed up & \textbf{1.21}  & \textbf{1.08}  & \textbf{1.57} & \textbf{1.31} \\
          & p-value & \textbf{0.163} & \textbf{0.371} & \textbf{0.005} & \textbf{0.048} \\
    \midrule
    \multirow{2}[2]{*}{CHOA, R=40} & Speed up & \textbf{1.38} & \textbf{1.29}  & \textbf{2.31} & \textbf{1.69} \\
          & p-value & \textbf{0.01} & \textbf{0.032} & \textbf{0.0005} & \textbf{0.002} \\
    \midrule
    \multirow{2}[2]{*}{CMS, R=15} & Speed up & \textbf{1.21} & 0.84  & \textbf{1.82} & \textbf{1.36} \\
          & p-value & \textbf{0.125} & 1.956 & \textbf{0.002} & \textbf{0.018} \\
    \midrule
    \multirow{2}[2]{*}{CMS, R=40} & Speed up & \textbf{1.57} & 0.87  & \textbf{2.51} & 0.99 \\
          & p-value & \textbf{0.00005} & 1.986 & \textbf{0.000004} & 1.08 \\
    \bottomrule
    \end{tabular}%
    }
  \label{tab:t-test}%
\end{table}%

\begin{figure}[ht]
    \centering
    \begin{subfigure}[t]{0.35\textwidth}
        \includegraphics[height=1.5in]{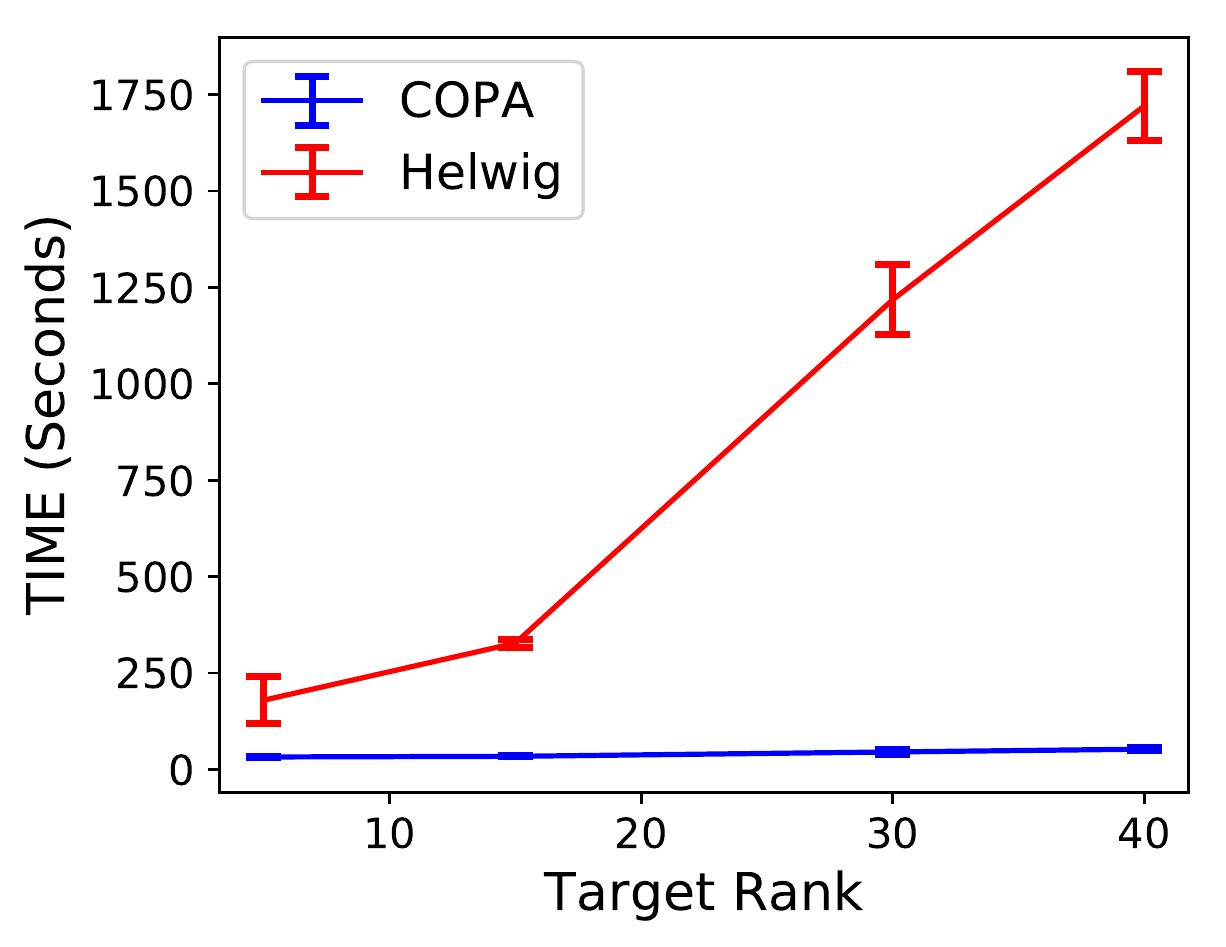}
        \caption{CHOA data set}
    \end{subfigure}%
    \begin{subfigure}[t]{0.35\textwidth}
        \includegraphics[height=1.5in]{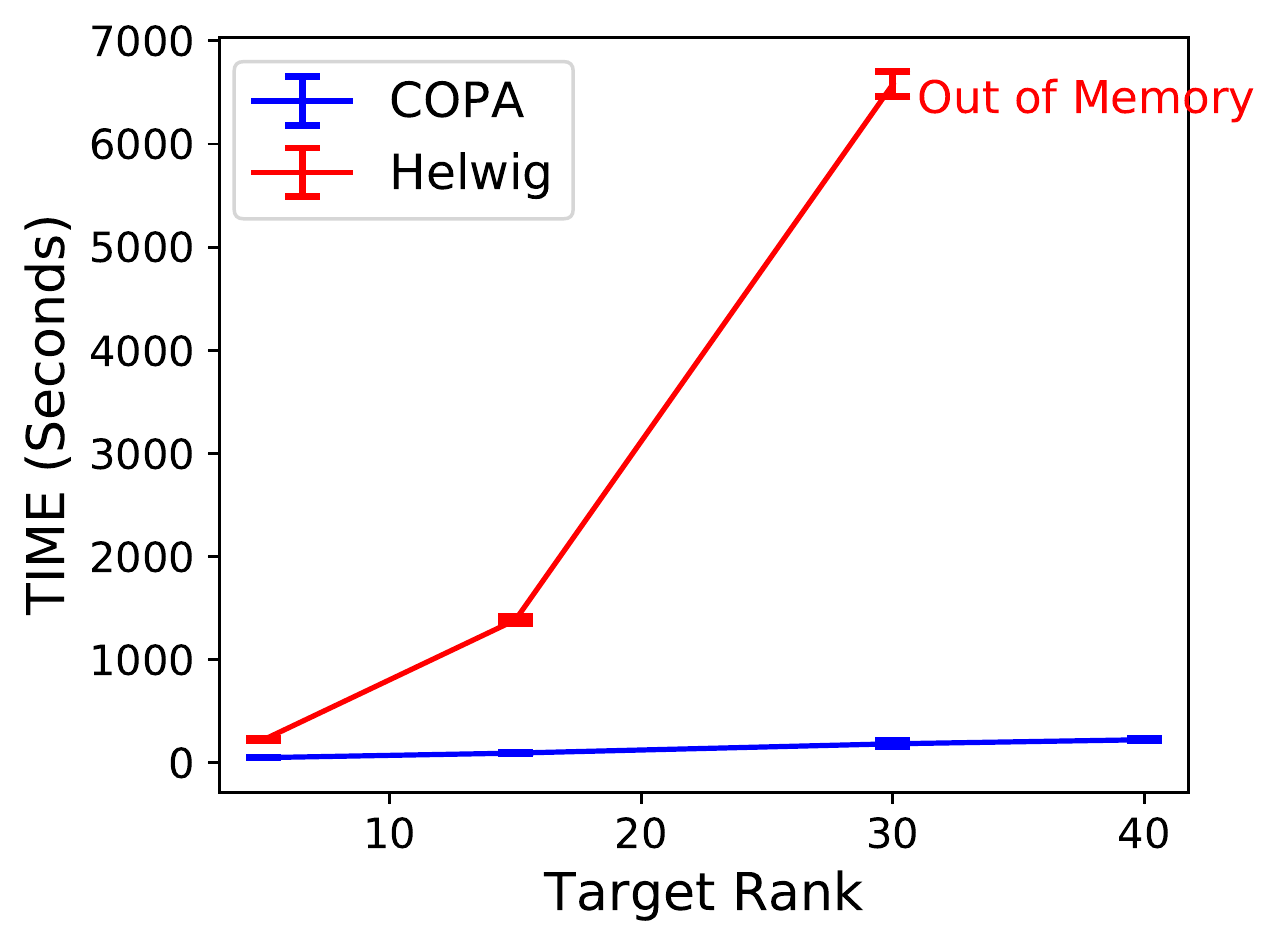}
        \caption{CMS data set.}
    \end{subfigure}
    \caption{\footnotesize Time in seconds for one iteration (as an average of 5 different random initializations) for different values of R. The left figure is the comparison on CHOA and the right figure shows the comparison on CMS. For R=40 \methodName~ achieves $32 \times$  over the Helwig approach on CHOA while for CMS dataset, execution in Helwig failed due to the excessive amount of memory request and \methodName finished an iteration with  the average of 224.21 seconds. }
    \label{fig:COPA_Helwig}
\end{figure}

In addition, we assessed the scalability of incorporating temporal smoothness onto {\small$\M{U_k}$} and compare it with Helwig's approach~\cite{helwig2017estimating} as SPARTan does not have the smoothness constraint.
Figure~\ref{fig:COPA_Helwig} provides a comparison of iteration time for Smooth \methodName~ and the approach in~\cite{helwig2017estimating} across two different target ranks.
First, we remark that our method is more scalable and faster than the baseline. For $R=30$, \methodName~is $27\times$ and $36\times$ faster on CHOA and CMS respectively. Moreover, for $R=40$, not only was \methodName $32\times$ faster on CHOA, but the execution failed using the approach in~\cite{helwig2017estimating} on CMS because of the excessive amount of memory required. In contrast, \methodName~successfully finished each iteration with an average of $224.21$ seconds.

\pgfplotstableread[col sep=space]{CHOA/SPARTan_15.txt}\datatable
\pgfplotstableread[col sep=space]{CHOA/AOADMM_15.txt}\datatable

\pgfplotstableread[col sep=space]{CHOA/SPARTan_40.txt}\datatable
\pgfplotstableread[col sep=space]{CHOA/AOADMM_40.txt}\datatable

\pgfplotstableread[col sep=space]{CMS/SPARTan_15.txt}\datatable
\pgfplotstableread[col sep=space]{CMS/AOADMM_15.txt}\datatable

\pgfplotstableread[col sep=space]{CMS/SPARTan_40.txt}\datatable
\pgfplotstableread[col sep=space]{CMS/AOADMM_40.txt}\datatable

\subsection{Case Study: CHOA Phenotype Discovery} \label{sec:pheno_disc}
\subsubsection{\textbf{Model interpretation:}}
Phenotyping is the process of extracting  a set of meaningful medical features  from raw and noisy EHRs. We define the following model interpretations regarding to our target case study: 
\begin{itemize}
\item Each column of factor matrix $\M{V}$ represents a phenotype and each row indicates a medical feature. Therefore an entry $\M{V}(i,j)$ represents the membership of medical feature $i$ to the $j^{th}$ phenotype.
\item The $r^{th}$ column of $\M{U_k} \in \mathbb{R}^{I_k\times R}$ indicates the evolution of  phenotype $r$ for all $I_k$ clinical visits for patient $k$.
\item The diagonal \M{S_k} provides the importance membership of R phenotypes for the  patient $k$. By sorting the values $diag(\M{S_k})$ we can identify the most important phenotypes for  patient $k$.
\end{itemize}
   
\subsubsection{\textbf{Case Study Setup:}} 
For this case study, we incorporate smoothness on {\small$\M{U_k}$}, non-negativity on {\small$\M{S_k}$}, and sparsity on {\small$\M{V}$} simultaneously to extract phenotypes from a subset of medically complex patients from CHOA dataset. These are the patients with high utilization, multiple specialty visits
and high severity. A total of $4602$ patients are selected with 810 distinct medical features. For this experiment, we set the number of basis functions to 7 (as shown in figure \ref{fig:M_bspline}), $\mu=49$, and $R=4$.
\subsubsection{\textbf{Findings:}} 
\begin{figure*}[htp]
\centering
\begin{subfigure}[b]{0.3\textwidth}
\centering
\includegraphics[width=.8\textwidth]{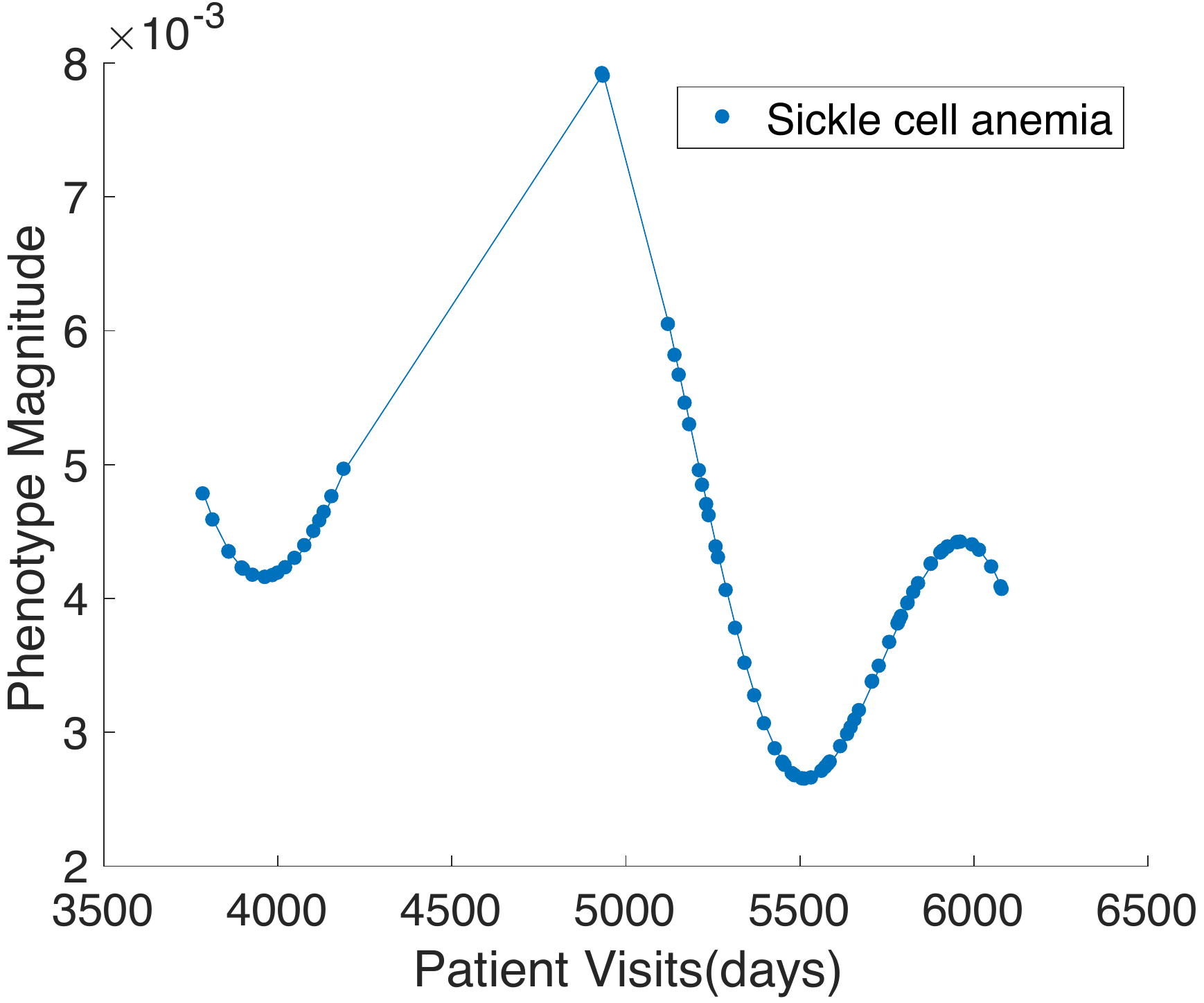}
\caption{ \methodName~}
\end{subfigure}%
\begin{subfigure}[b]{0.3\textwidth}
\centering
\includegraphics[width=.8\textwidth]{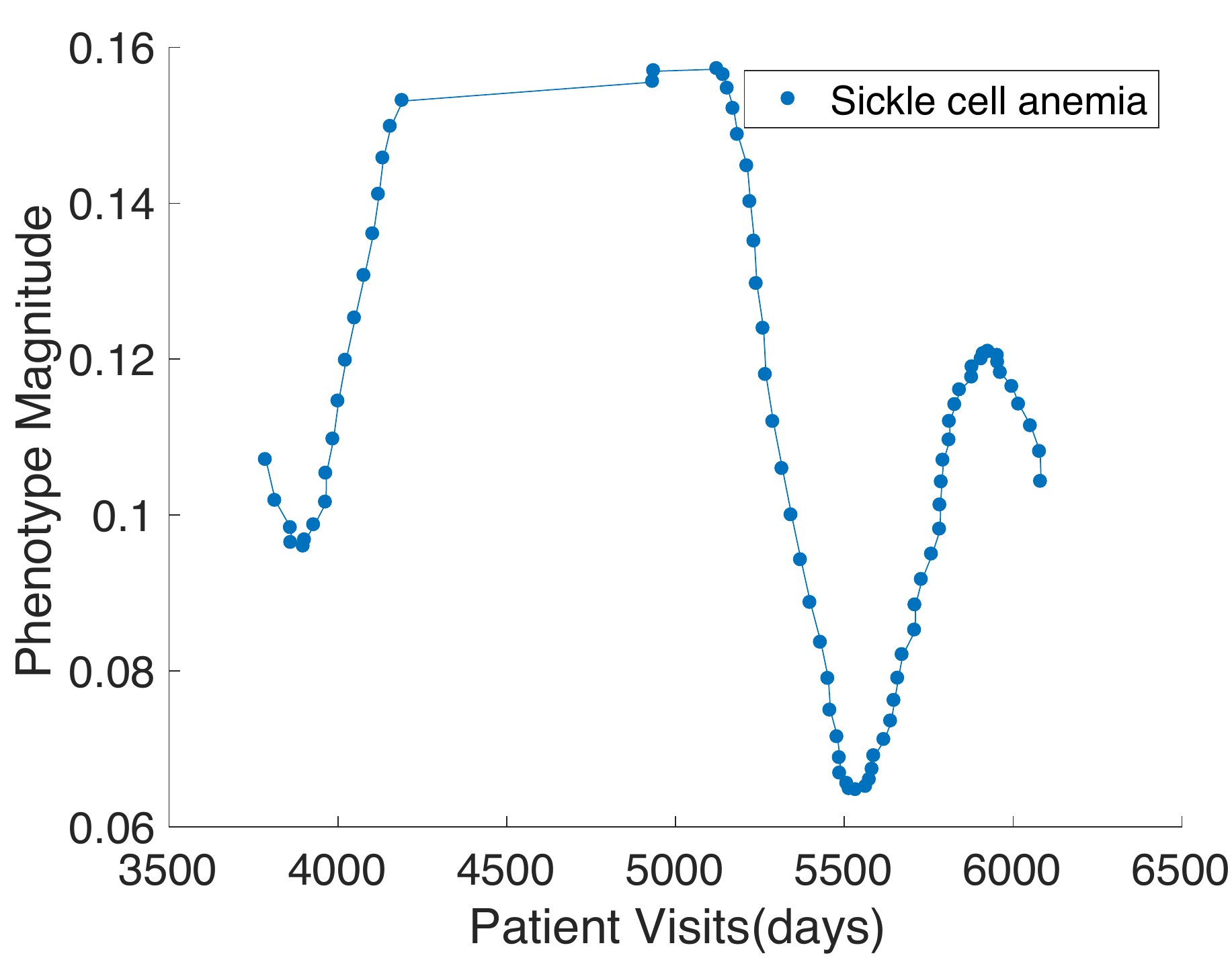}
\caption{Helwig}
\end{subfigure}%
\begin{subfigure}[b]{0.3\textwidth}
\centering
\includegraphics[width=.8\textwidth]{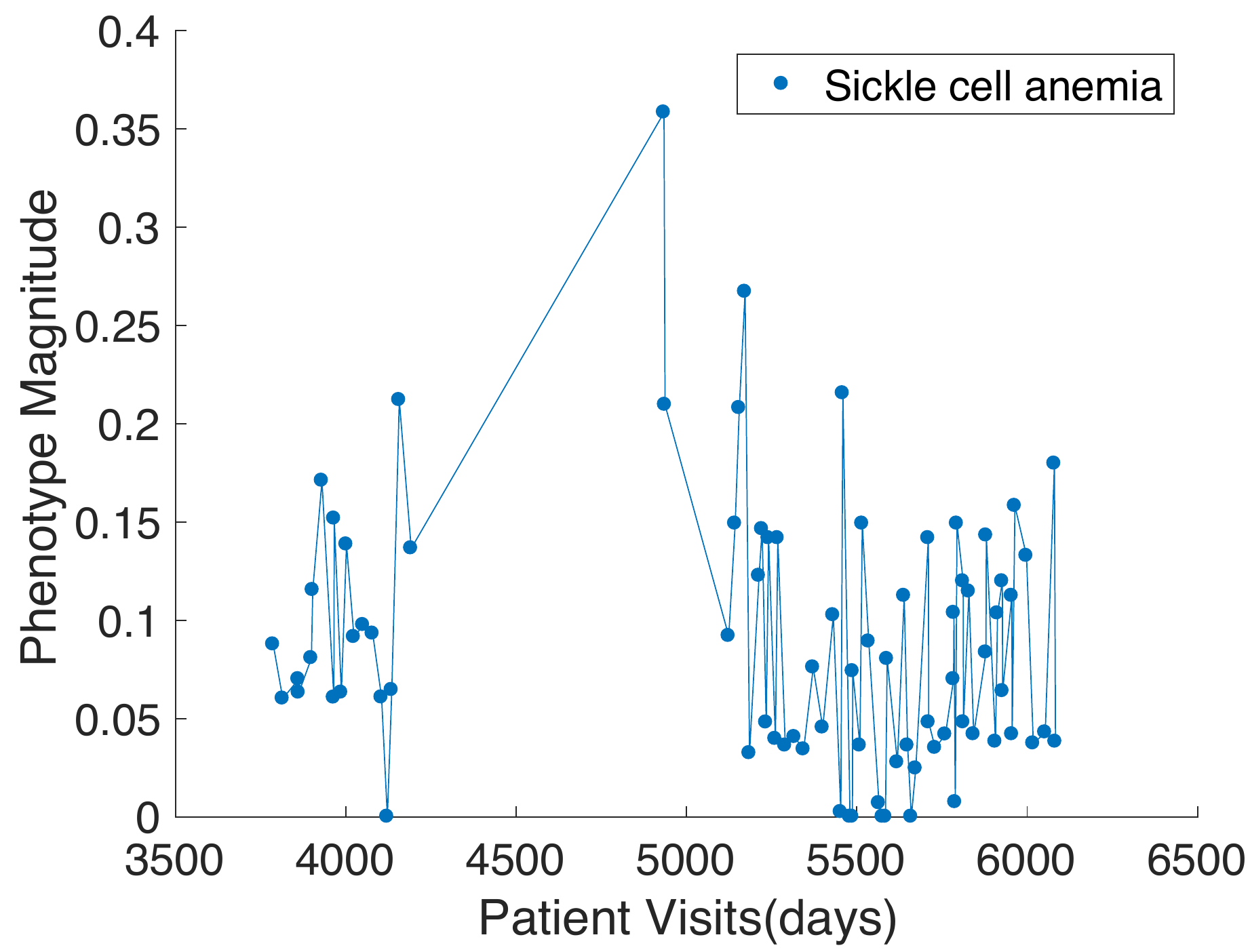}
\caption{SPARTan}
\end{subfigure}
\begin{subfigure}[b]{0.3\textwidth}
\centering
\includegraphics[width=.8\textwidth]{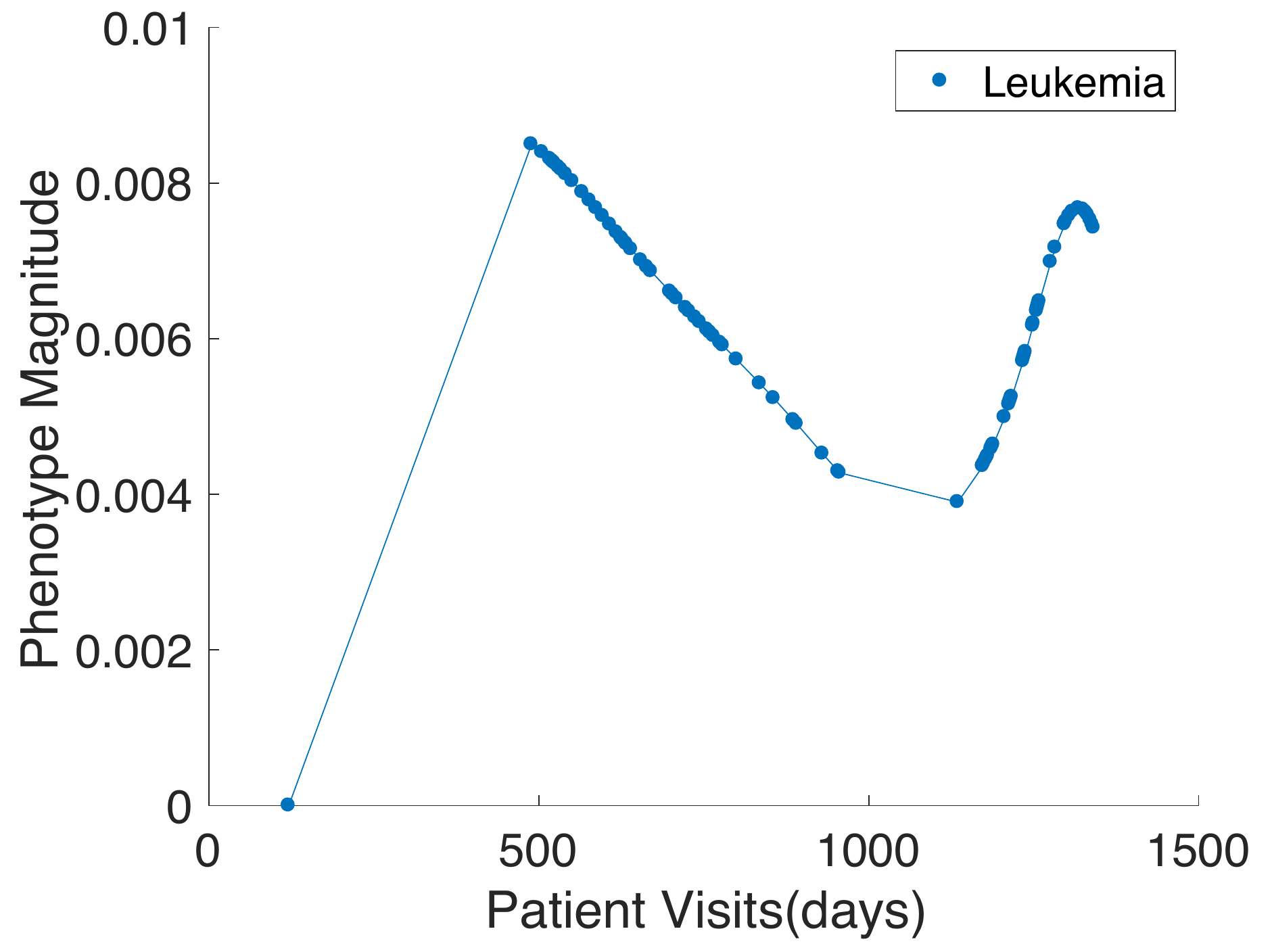}
\caption{ \methodName~}
\end{subfigure}%
\begin{subfigure}[b]{0.3\textwidth}
\centering
\includegraphics[width=.8\textwidth]{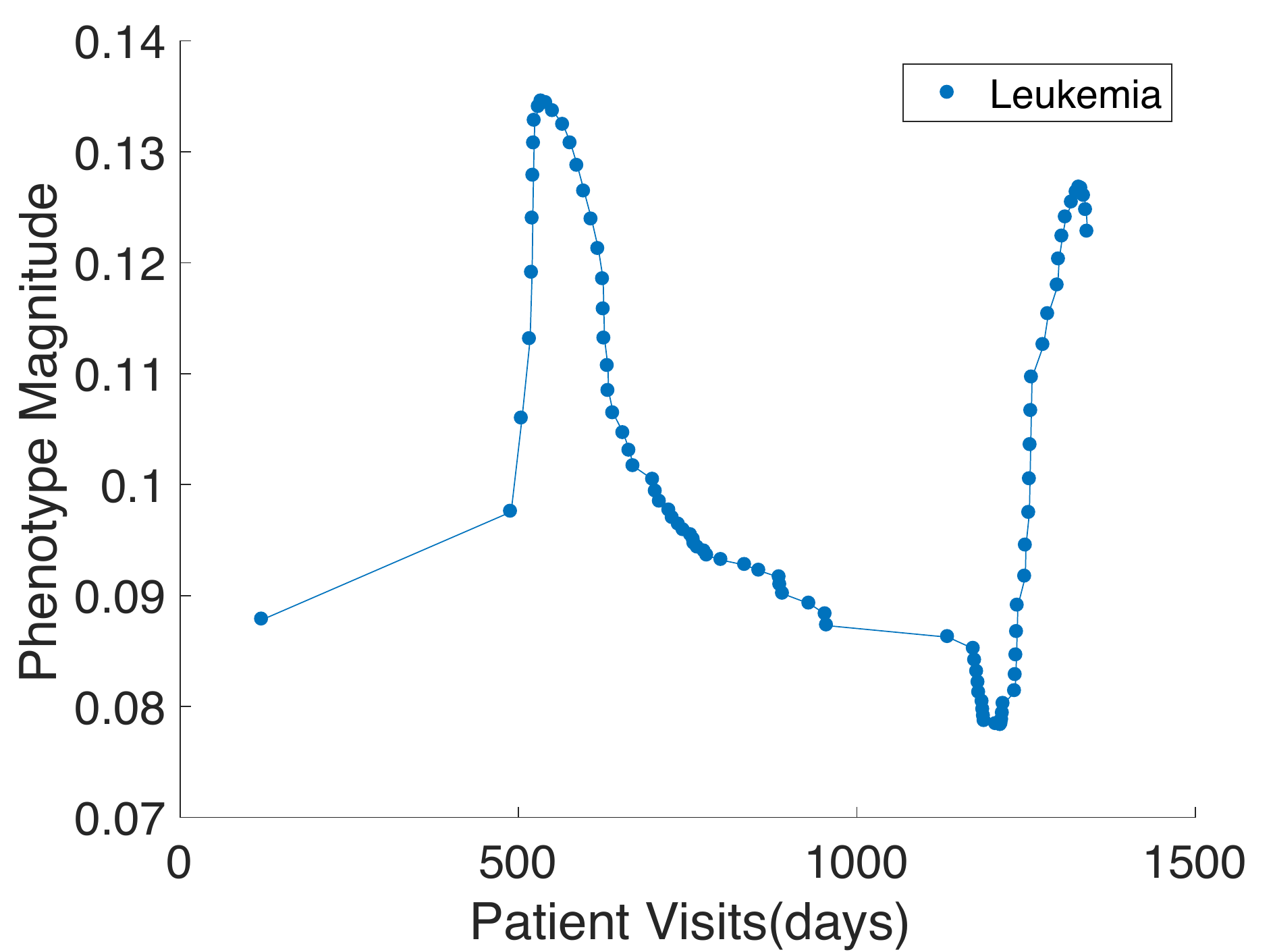}
\caption{Helwig}
\end{subfigure}%
\begin{subfigure}[b]{0.3\textwidth}
\centering
\includegraphics[width=.8\textwidth]{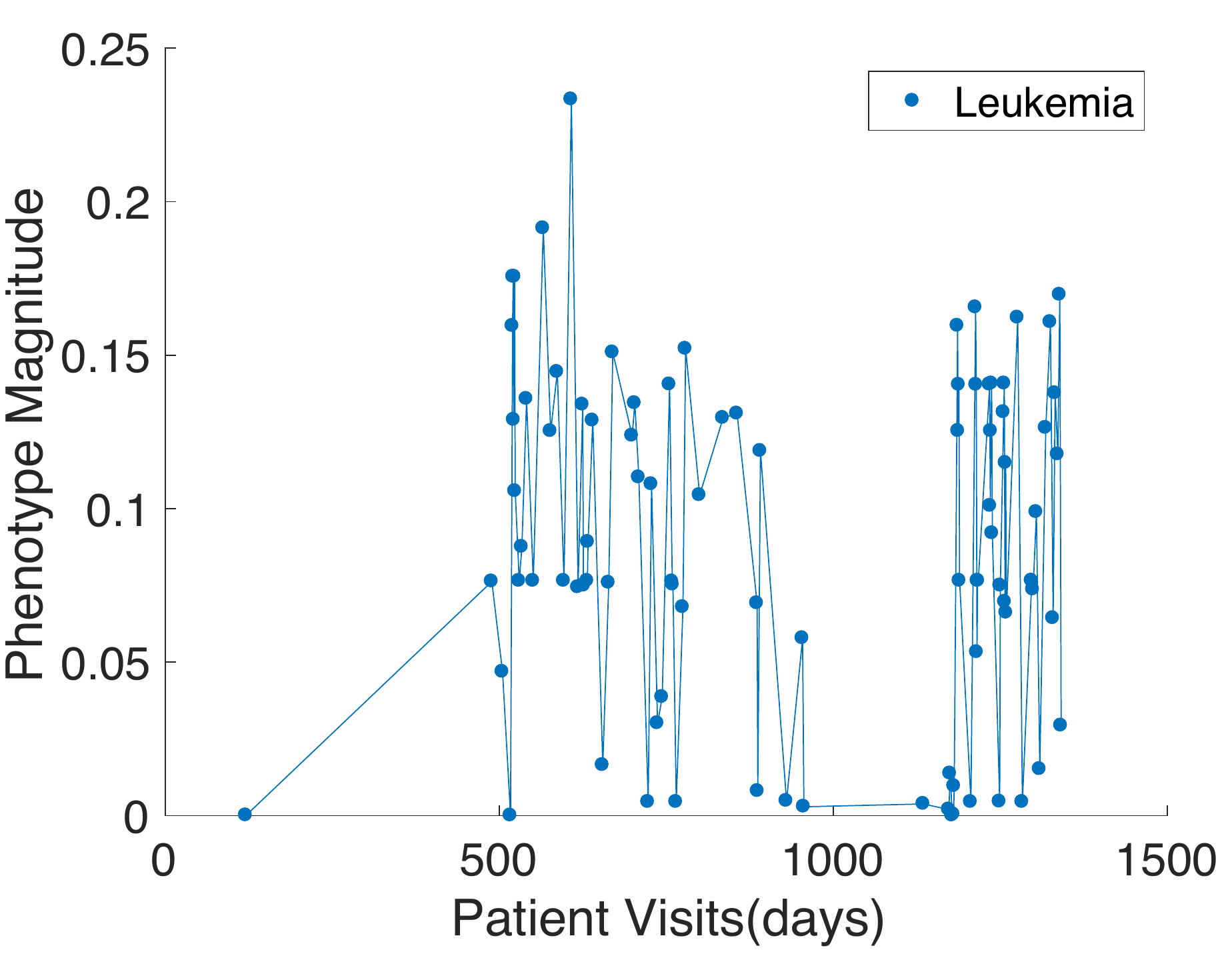}
\caption{SPARTan}
\end{subfigure}
\caption{\footnotesize The temporal patterns extracted for two patients by \methodName~, Helwig, and SPARTan. The first row is associated with a patient who has sickle cell anemia while the second row is for a patient with Leukemia.}
\label{fig:temporal_signature}
\end{figure*}

We demonstrate the effectiveness of \methodName~for extracting phenotypes. Also we show how \methodName is able to describe the evolution of phenotypes for patients by considering the gap between every pair of clinical visits. 
Figure \ref{fig:temporal_signature} displays the evolution of phenotypes (temporal pattern) relating to two patients discovered by \methodName, Helwig, and SPARTan. The phenotype that is chosen has the highest weight for each patient (largest value in the diagonal {\small$\M{S_k}$} matrix) and the loadings on the medical features are similar across all three methods. The first row in figure \ref{fig:temporal_signature} is from a patient who has sickle cell anemia. There is a large gap between the $19^{th}$ and $20^{th}$ visits (742 days or $\sim 2$ years) with a significant increase in the occurrence of medications/diagnosis in the patient's EHR record. \methodName~ models this difference and yields phenotype loadings that capture this drastic change. 
On the other hand, the factor resulting from Helwig's approach assumes the visits are close in time and produce the same magnitude for the next visit. The second row in figure \ref{fig:temporal_signature} reflects the temporal signature for a patient with Leukemia. In the patient's EHRs, the first visit occurred on day 121 without any sign of Leukemia. The subsequent visit (368 days later) reflects a change in the patient's status with a large number of diagnosis and medications. \methodName~ encapsulates this phenomenon, while the Helwig factor suggests the presence of Leukemia at the first visit which is not present. Although SPARTan produces temporally-evolving phenotypes, it treats time as a categorical feature. Thus, there are sudden spikes in the temporal pattern which hinders interpretability and clinical meaningfulness. 

Next, we present the phenotypes discovered by \methodName in table \ref{tab:phenotypes}. It is important to note that no additional post-processing was performed on these results. These four phenotypes have been endorsed by a clinical expert as clinically meaningful. Moreover, the expert has provided the labels to reflect the associated medical concept. As the phenotypes discovered by SPARTan and Helwig are too dense and require significant post-processing, they are not displayed in this paper.

\begin{table}[htbp]
  \centering
  \caption{\footnotesize  Phenotypes discovered by \methodName~. The red color corresponds to diagnosis and blue color corresponds to medication. The meaningfulness of phenotypes endorsed by a medical expert. No additional post-processing was performed on these results.}
   \scalebox{0.8}{
    \begin{tabular}{l}
    \textcolor[rgb]{ .133,  .133,  .133}{\textbf{Leukemias}} \\
    \midrule
    \textcolor[rgb]{ 1,  0,  0}{Leukemias} \\
    \textcolor[rgb]{ 1,  0,  0}{Immunity disorders} \\
    \textcolor[rgb]{ 1,  0,  0}{Deficiency and other anemia} \\
    \textcolor[rgb]{ .067,  .333,  .8}{HEPARIN AND RELATED PREPARATIONS} \\
    \textcolor[rgb]{ .067,  .333,  .8}{Maintenance chemotherapy; radiotherapy} \\
    \textcolor[rgb]{ .067,  .333,  .8}{ANTIEMETIC/ANTIVERTIGO AGENTS} \\
    \textcolor[rgb]{ .067,  .333,  .8}{SODIUM/SALINE PREPARATIONS} \\
    \textcolor[rgb]{ .067,  .333,  .8}{TOPICAL LOCAL ANESTHETICS} \\
    \textcolor[rgb]{ .067,  .333,  .8}{GENERAL ANESTHETICS INJECTABLE} \\
    \textcolor[rgb]{ .067,  .333,  .8}{ANTINEOPLASTIC - ANTIMETABOLITES} \\
    \textcolor[rgb]{ .067,  .333,  .8}{ANTIHISTAMINES - 1ST GENERATION} \\
    \textcolor[rgb]{ .067,  .333,  .8}{ANALGESIC/ANTIPYRETICS NON-SALICYLATE} \\
    \textcolor[rgb]{ .067,  .333,  .8}{ANALGESICS NARCOTIC ANESTHETIC ADJUNCT AGENTS} \\
    \textcolor[rgb]{ .067,  .333,  .8}{ABSORBABLE SULFONAMIDE ANTIBACTERIAL AGENTS} \\
    \textcolor[rgb]{ .067,  .333,  .8}{GLUCOCORTICOIDS} \\
    \midrule
    \textcolor[rgb]{ .133,  .133,  .133}{\textbf{Neurological Disorders}} \\
    \midrule
    \textcolor[rgb]{ 1,  0,  0}{Other nervous system disorders} \\
    \textcolor[rgb]{ 1,  0,  0}{Epilepsy; convulsions} \\
    \textcolor[rgb]{ 1,  0,  0}{Paralysis} \\
    \textcolor[rgb]{ 1,  0,  0}{Other connective tissue disease} \\
    \textcolor[rgb]{ 1,  0,  0}{Developmental disorders} \\
    \midrule
    \textcolor[rgb]{ .067,  .333,  .8}{Rehabilitation care; and adjustment of devices} \\
    \textcolor[rgb]{ .067,  .333,  .8}{ANTICONVULSANTS} \\
    \midrule
    \textcolor[rgb]{ .133,  .133,  .133}{\textbf{Congenital anomalies}} \\
    \midrule
    \textcolor[rgb]{ 1,  0,  0}{Other perinatal conditions} \\
    \textcolor[rgb]{ 1,  0,  0}{Cardiac and circulatory congenital anomalies} \\
    \textcolor[rgb]{ 1,  0,  0}{Short gestation; low birth weight} \\
    \textcolor[rgb]{ 1,  0,  0}{Other congenital anomalies} \\
    \midrule
    \textcolor[rgb]{ 0,  .439,  .753}{Fluid and electrolyte disorders} \\
    \textcolor[rgb]{ 0,  .439,  .753}{LOOP DIURETICS} \\
    \textcolor[rgb]{ 0,  .439,  .753}{IV FAT EMULSIONS} \\
    \midrule
    \textcolor[rgb]{ .133,  .133,  .133}{\textbf{Sickle Cell Anemia}} \\
    \midrule
    \textcolor[rgb]{ 1,  0,  0}{Sickle cell anemia} \\
    \textcolor[rgb]{ 1,  0,  0}{Other gastrointestinal disorders} \\
    \textcolor[rgb]{ 1,  0,  0}{Other nutritional; endocrine; and metabolic disorders} \\
    \textcolor[rgb]{ 1,  0,  0}{Other lower respiratory disease} \\
    \textcolor[rgb]{ 1,  0,  0}{Asthma} \\
    \textcolor[rgb]{ 1,  0,  0}{Allergic reactions} \\
    \textcolor[rgb]{ 1,  0,  0}{Esophageal disorders} \\
    \textcolor[rgb]{ 1,  0,  0}{Respiratory failure; insufficiency; arrest (adult)} \\
    \textcolor[rgb]{ 1,  0,  0}{Other upper respiratory disease} \\
    \textcolor[rgb]{ .067,  .333,  .8}{BETA-ADRENERGIC AGENTS} \\
    \textcolor[rgb]{ .067,  .333,  .8}{ANALGESICS NARCOTICS} \\
    \textcolor[rgb]{ .067,  .333,  .8}{NSAIDS, CYCLOOXYGENASE INHIBITOR - TYPE} \\
    \textcolor[rgb]{ .067,  .333,  .8}{ANALGESIC/ANTIPYRETICS NON-SALICYLATE} \\
    \textcolor[rgb]{ .067,  .333,  .8}{POTASSIUM REPLACEMENT} \\
    \textcolor[rgb]{ .067,  .333,  .8}{SODIUM/SALINE PREPARATIONS} \\
    \textcolor[rgb]{ .067,  .333,  .8}{GENERAL INHALATION AGENTS} \\
    \textcolor[rgb]{ .067,  .333,  .8}{LAXATIVES AND CATHARTICS} \\
    \textcolor[rgb]{ .067,  .333,  .8}{IV SOLUTIONS: DEXTROSE-SALINE} \\
    \textcolor[rgb]{ .067,  .333,  .8}{ANTIEMETIC/ANTIVERTIGO AGENTS} \\
    \textcolor[rgb]{ .067,  .333,  .8}{SEDATIVE-HYPNOTICS NON-BARBITURATE} \\
    \textcolor[rgb]{ .067,  .333,  .8}{GLUCOCORTICOIDS, ORALLY INHALED} \\
    \textcolor[rgb]{ .067,  .333,  .8}{FOLIC ACID PREPARATIONS} \\
    \textcolor[rgb]{ .067,  .333,  .8}{ANALGESICS NARCOTIC ANESTHETIC ADJUNCT AGENTS} \\
    \bottomrule
    \end{tabular}%
    }
 \label{tab:phenotypes}%
\end{table}%

\section{Related Work}
SPARTan was proposed for PARAFAC2 modeling on large and sparse data \cite{Perros2017-dh}. A specialized Matricized-Tensor-Times-Khatri-Rao-Product (MTTKRP) was designed to efficiently decompose the tensor {\small$\T{Y}$} ({\small$\M{Y_k} = \M{Q_k^T} \M{X_k}$}) both in terms of speed and memory. Experimental results demonstrate the scalability of this approach for large and sparse datasets. However, the target model and the fitting algorithm do not enable imposing constraints such as smoothness and sparsity, which would enhance the interpretability of the model results.
A small number of works have introduced constraints (other than non-negativity) for the PARAFAC2 model.
  Helwig~\cite{helwig2017estimating} imposed both functional and structural constraints. Smoothness (functional constraint) was incorporated by extending the use of basis functions introduced for CP \cite{timmerman2002three}. Structural information (variable loadings) were formulated using Lagrange multipliers \cite{clarke1976new} by modifying the CP-ALS algorithm. Unfortunately, Helwig's algorithm suffers the same computational and memory bottlenecks as the classical algorithm designed for dense data~\cite{kiers1999parafac2}. Moreover, the formulation does not allow for easy extensions of other types of constraints (e.g., sparsity).

Other works that tackle the problem of computational phenotyping through constrained tensor factorization (e.g., \cite{ho2014marble,wang2015rubik}) cannot handle irregular tensor input (as summarized in Table~\ref{propterty_comp}); thus they are limited to aggregating events across time, which may lose temporal patterns providing useful insights. 

\section{Conclusion}
Interpretable and meaningful tensor factorization models are desirable.
One way to improve the interpretability of tensor factorization approaches is by introducing constraints such as sparsity, non-negativity, and smoothness.
However, existing constrained tensor factorization methods are not well-suited for an irregular tensor.
While PARAFAC2 is a suitable model for such data, there is no general and scalable framework for imposing constraints in PARAFAC2.

Therefore, in this paper we propose, \methodName, a constrained PARAFAC2 framework for large and sparse data. Our framework is able to impose constraints simultaneously by applying element-wise operations. Our motivating application is extracting temporal patterns and phenotypes from noisy and raw EHRs.
By incorporating smoothness and sparsity, we produce meaningful phenotypes and patient temporal signatures that are confirmed by a clinical expert. 
\section{ACKNOWLEDGMENT}
This work was supported by the National Science Foundation, award IIS-$\#$1418511 and CCF-$\#$1533768, the National Institute of Health award 1R01MD011682-01 and R56HL138415, Children's Healthcare of Atlanta, and the National Institute of Health under award number 1K01LM012924-01. Research at UCR was supported by the Department of the Navy, Naval Engineering Education Consortium under award no. N00174-17-1-0005 and by an Adobe Data Science Research Faculty Award.

\bibliographystyle{unsrt}
\bibliography{main.bib} 
\end{document}